\documentclass[10pt,journal,compsoc]{IEEEtran}

\ifCLASSOPTIONcompsoc
\usepackage[nocompress]{cite}
\else
\usepackage{cite}
\fi
\usepackage{amsmath}
\usepackage{amssymb}
\usepackage{amsthm}
\usepackage{multirow}
\usepackage{array}
\usepackage{bm}
\usepackage{enumerate}

\usepackage{booktabs}
\usepackage{graphicx}
\usepackage{adjustbox}
\usepackage{relsize}

\usepackage{algorithmic}
\usepackage{algorithm}

\usepackage{url}
\usepackage{bbm}
\usepackage{dsfont}
\usepackage{array}
\usepackage{courier}
\usepackage{bbding}

\usepackage{color}
\usepackage{subfigure}

\usepackage{rotating}
\usepackage{orcidlink}

\usepackage{makecell}

\begin{document}
	%
	
	\title{Calibrated One-class Classification for \\ Unsupervised Time Series Anomaly Detection}

	%
	%
	
	\author{Hongzuo~Xu\orcidlink{0000-0001-8074-1244},
		Yijie~Wang\orcidlink{0000-0002-2913-4016},
		Songlei~Jian\orcidlink{0000-0002-1435-0410},
		Qing~Liao\orcidlink{0000-0003-1012-5301},
		Yongjun~Wang\orcidlink{0000-0002-3257-539X} and
		Guansong~Pang\orcidlink{0000-0002-9877-2716}
		\IEEEcompsocitemizethanks{
			\IEEEcompsocthanksitem 
			Hongzuo Xu is with the Intelligent Game and Decision Lab (IGDL), Beijing 100091, PR China.
			(e-mail: hongzuoxu@126.com) 
			\IEEEcompsocthanksitem 
			Yijie Wang is with the National Key Laboratory of Parallel and Distributed Computing, College of Computer, National University of Defense Technology, Hunan 410073, PR China. (e-mail: wangyijie@nudt.edu.cn)
			\IEEEcompsocthanksitem 
			Songlei Jian and Yongjun Wang are with the College of Computer, National University of Defense Technology, Hunan 410073, PR China. 
			(e-mail: \{jiansonglei, wangyongjun\}@ nudt.edu.cn)
			\IEEEcompsocthanksitem Qing Liao is with the School of Computer Science and Technology, Harbin Institute of Technology (Shenzhen), Guangdong 518055, PR China. (e-mail: liaoqing@hit.edu.cn)
			\IEEEcompsocthanksitem Guansong Pang is with the School of Computing and Information Systems, Singapore Management University Singapore, 178902, Singapore. (e-mail: gspang@smu.edu.sg)
			\IEEEcompsocthanksitem Yijie Wang is the corresponding author. 
		}
	}

\IEEEtitleabstractindextext{%
\begin{abstract}
Time series anomaly detection is instrumental in maintaining system availability in various domains. Current work in this research line mainly focuses on learning data normality deeply and comprehensively by devising advanced neural network structures and new reconstruction/prediction learning objectives. However, their one-class learning process can be misled by latent anomalies in training data (i.e., anomaly contamination) under the unsupervised paradigm. Their learning process also lacks knowledge about the anomalies. Consequently, they often learn a biased, inaccurate normality boundary. To tackle these problems, this paper proposes calibrated one-class classification for anomaly detection, realizing contamination-tolerant, anomaly-informed learning of data normality via uncertainty modeling-based calibration and native anomaly-based calibration. Specifically, our approach adaptively penalizes uncertain predictions to restrain irregular samples in anomaly contamination during optimization, while simultaneously encouraging confident predictions on regular samples to ensure effective normality learning. This largely alleviates the negative impact of anomaly contamination. Our approach also creates native anomaly examples via perturbation to simulate time series abnormal behaviors. Through discriminating these dummy anomalies, our one-class learning is further calibrated to form a more precise normality boundary. Extensive experiments on ten real-world datasets show that our model achieves substantial improvement over sixteen state-of-the-art contenders.



\end{abstract}

}

\maketitle

\IEEEdisplaynontitleabstractindextext

%
\IEEEpeerreviewmaketitle

\IEEEraisesectionheading{\section{Introduction}\label{sec:introduction}}

%
%
%
%
\IEEEPARstart{O}{ver} recent decades, with the burgeoning of informatization, a substantial amount of time series data have been created. 
As the functioning status of various target systems such as large-scale data centers, cloud servers, and space crafts, these time series data are a source where we can monitor and alarm potential faults, threats, and risks of target systems by identifying their unusual states (i.e., anomalies).
Anomaly detection, an important field in data mining and analytics, is to find exceptional data observations that deviate significantly from the majority \cite{pang2021deep}, which plays a critical role in achieving this goal.
Due to the cost and difficulty of labeling work in these real-world applications, time series anomaly detection is often formulated as an unsupervised task with unlabeled training data.

\begin{figure}[t]
	\centering
	\includegraphics[width=0.46\textwidth]{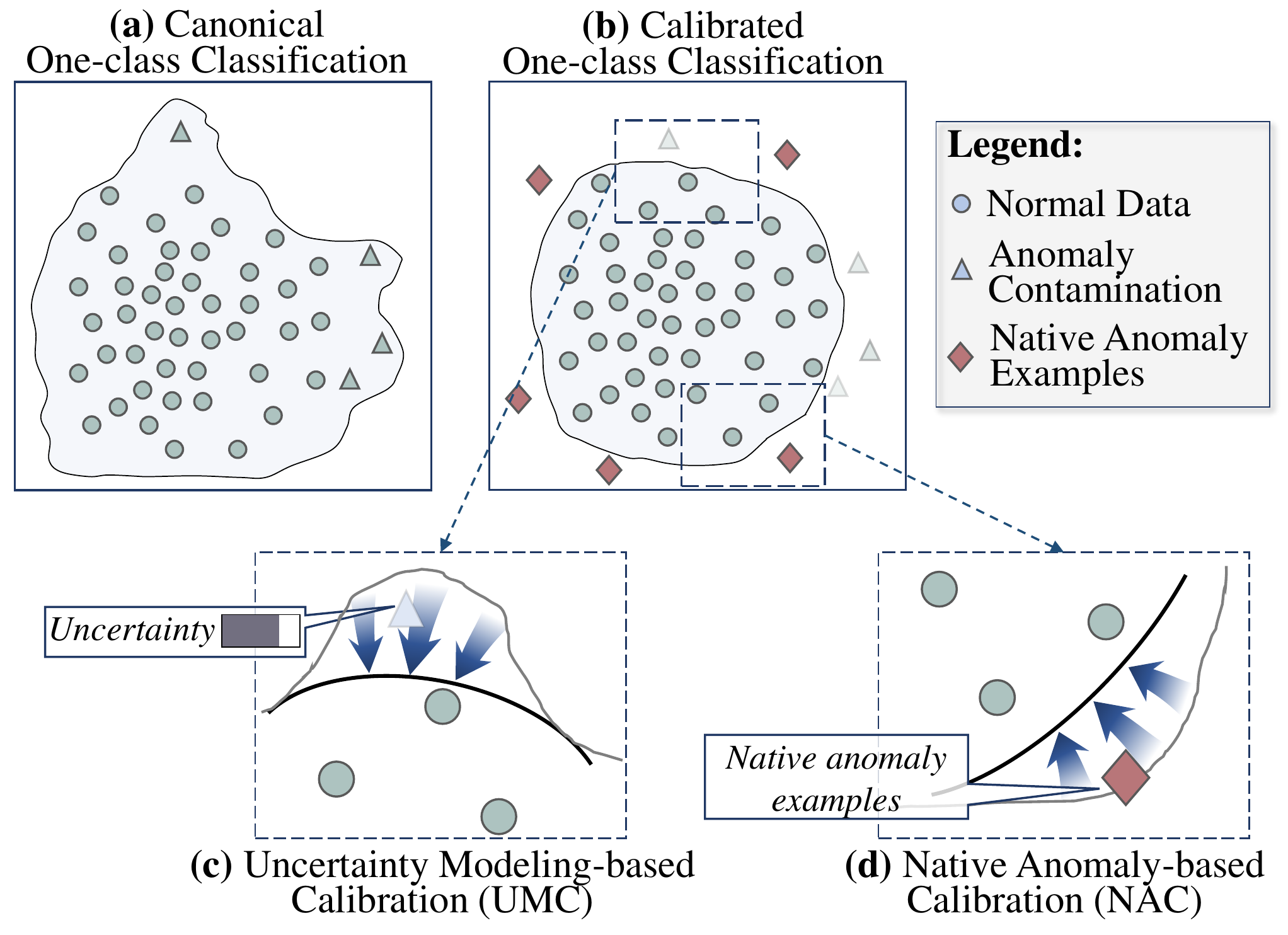}
	\caption{
		Demonstration of our key insights. 
		\textbf{(a)} Broadly-used canonical one-class classification 
		may learn an inaccurate, biased normality boundary due to the anomaly contamination problem and the absence of knowledge about anomalies.
		\textbf{(b)} By contrast, the two proposed calibration methods, UMC and NAC, respectively address these two issues, and our calibrated one-class classifier can produce a more accurate, clearer normality boundary. 
		\textbf{(c)} UMC helps weaken contaminated data during optimization based on model uncertainty scores, while (\textbf{d}) NAC helps ground the normality boundary by calibrating the normality with native anomaly examples.  
	}
	\label{fig:demonstration}
\end{figure}

\textbf{Challenges.} 	
Without any guidance of supervisory signals, unsupervised time series anomaly detection generally relies on modeling data normality via one-class learning because most training samples are normal. 
However, this learning process still faces two key challenges: (1) \textit{the presence of anomalies in training sets}, and (2) \textit{the absence of knowledge about the anomalies}.
Specifically, as has been done in mainstream methods like \cite{deng2021gdn,tuli2022tranad,audibert2020usad,xu2022transformer}, 
the whole training set is often directly fed into learning models by assuming all training samples are normal.
However, this assumption does not always hold. The training set might not be pure but contaminated by a small part of anomalies.
Anomaly contamination can greatly disturb the learning process, thus leading to severe overfitting problems.
Besides, the learning process, without any knowledge about anomalies, may find an inaccurate normality boundary, since it is hard to define the range of normal behaviors in such cases. As shown in Fig. \ref{fig:demonstration} (a), due to these two problems, canonical one-class classification methods typically learn a suboptimal normality model.

\textbf{Prior Art.} 
Most unsupervised time series anomaly detectors use generative models in one-class learning to restore input data \cite{tuli2022tranad,campos2021vldb,audibert2020usad,li2021multivariate} or forecast future data \cite{deng2021gdn,hundman2018detecting,zhao2020gat}. Data normality is implicitly learned behind the rationale of reconstruction or prediction. The abnormal degrees of observations in time series can be measured according to loss values. To achieve a comprehensive delineation of data normality (e.g., deeper inter-metric correlations, longer-term temporal dependence, and more diverse patterns), these methods design advanced network structures (e.g., using variational Autoencoders \cite{li2021multivariate,su2019omni}, graph neural networks \cite{deng2021gdn,zhao2020gat}, and Transformer \cite{xu2022transformer,tuli2022tranad}) and new reconstruction/prediction learning objectives (e.g., adding adversarial training \cite{liu2022tkde,kieu2022robust,audibert2020usad}, ensemble learning \cite{kieu2019ensemble,campos2021vldb}, and meta-learning \cite{tuli2022tranad}). 
However, these current methods generally do not have components to deal with the anomaly contamination issue.
There are a few attempts to address this problem, e.g., using pseudo-labels via self-training \cite{du2021tkde,pang2018learning,pang2020self} or an extra pre-positive one-class classification model \cite{kieu2022robust} to remove plausible anomalies in the training set. Nonetheless, these additional components themselves might be biased by the anomaly contamination, leading to high errors in the pseudo labeling or anomaly removal. They may also wrongly discard some normal samples of the boundary that are informative and vital in learning data normality.  
On the other hand, these methods do not consider information related to the anomaly class when performing their normality learning process. It is difficult to learn accurate normality without knowing what the abnormalities are.

\textbf{New Insights.} 
To address these challenges, this paper 
investigates an intriguing question: \textit{Can we calibrate one-class classification from two facets, i.e., alleviating the negative impact of anomaly contamination and introducing knowledge about anomalies,
to learn a contamination-tolerant, anomaly-informed data normality?}

As for the first facet, the essence is to eliminate the contribution of these noisy samples in the learning process.
We resort to model uncertainty to tackle this problem. These anomalies are typically accompanied by rare and inconsistent behaviors, and as a result, the one-class learning model tends to make predictions unconfidently. As shown in Fig. \ref{fig:demonstration} (c), we aim to use this type of uncertainty to weaken the contribution of anomaly contamination, thereby calibrating the one-class model w.r.t. the contaminated training data. 
Particularly, we can design a new learning objective embedded with the uncertainty concept. It adaptively penalizes uncertain predictions, while simultaneously encouraging more confident predictions to ensure effective learning of hard samples. Therefore, this process can discriminate these harmful anomalies in learning data normality, thus masking the notorious anomaly contamination problem during network optimization. 


To address the second facet, we are motivated by self-supervised learning that creates supervisory signals from the data itself. Current self-supervised anomaly detection methods design various supervised proxy tasks, e.g., geometric transformation prediction \cite{golan2018dagmm}, masking prediction \cite{li2021cutpaste}, and continuity identification \cite{deldari2021time}, but these tasks mainly contribute to learning clearer semantics of the input data rather than introducing information related to anomalies. 
Since time series anomalies are well characterized and defined as point, contextual, and collective anomalies \cite{lai2021revisiting}, we aim to utilize these definitions and characterizations to simulate abnormal behaviors by tailored data perturbation operations upon original time series data.
This process can offer reliable primitive anomaly examples, or at least data samples with abnormal semantics, to the one-class learning process.
As shown in Fig. \ref{fig:demonstration} (d), these \textit{native anomaly examples} (``native'' indicates that they are generated based on original data) can further help calibrate the discriminability of the learned normality.

\begin{figure*}[htbp]
	\centering 
	\subfigbottomskip=-10pt 
	\subfigcapskip=-5pt 
	\subfigure[\textbf{Time Series Data}]{
		\includegraphics[width=0.475\linewidth]{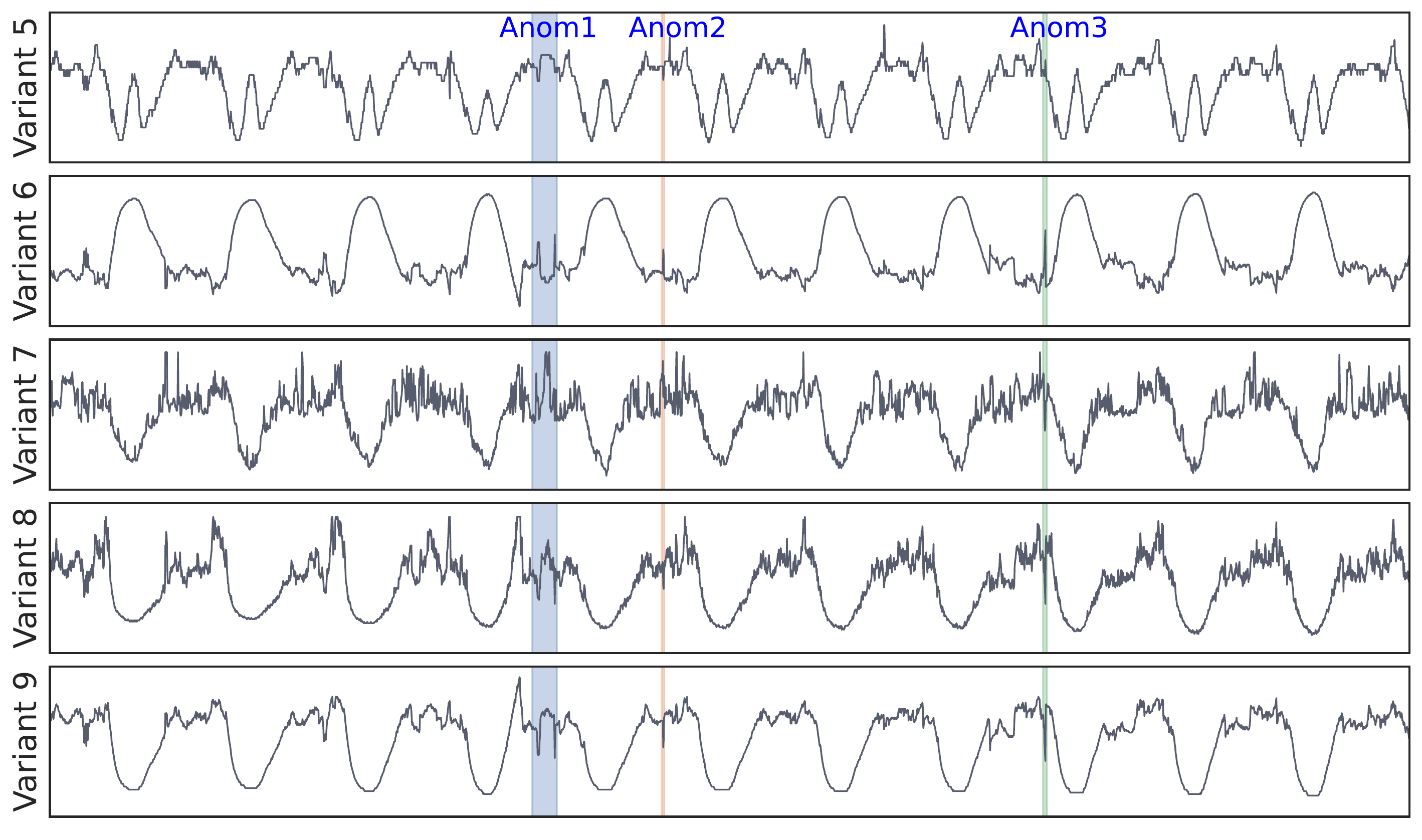}
	} 
	\hspace{0.4cm}
	\subfigure[\textbf{Feature Space}]{
		\includegraphics[width=0.376\linewidth]{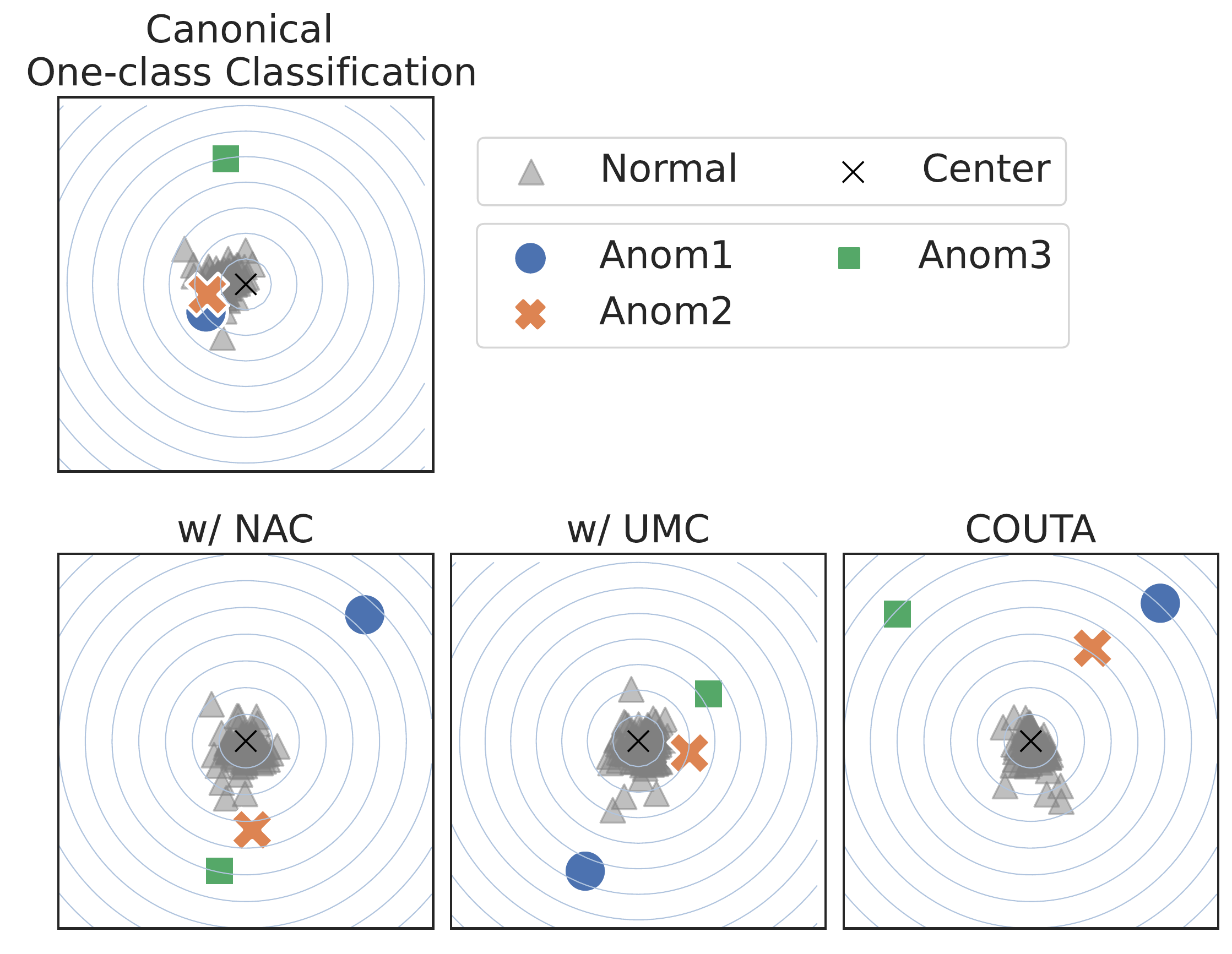}
	}
\caption{
\textbf{(a)} Time series data with three anomaly segments; \textbf{(b)} Learned feature space of canonical (non-calibrated) one-class classification and the proposed methods by using NAC/UMC separately and two calibration methods simultaneously (i.e., COUTA). Normal data is expected to be enclosed in a compact hypersphere, and anomalies can be successfully identified if they are distant from the center. }
	\label{fig:example}
\end{figure*}

\textbf{The Proposed Approach.} Based on the above motivation and insights, this paper proposes a novel Calibrated One-class classification-based Unsupervised Time series Anomaly detection method (COUTA for short). The approach fulfills contamination-tolerant, anomaly-informed normality learning by two novel normality calibration methods, including \textit{uncertainty modeling-based calibration (UMC)} and \textit{native anomaly-based calibration (NAC)}. 
In UMC, a novel calibrated one-class loss function is proposed, in which a prior (i.e., Gaussian) distribution is imposed on the one-class distances and utilized to capture the prediction uncertainties. It calibrates one-class representations by adaptively penalizing uncertain predictions while at the same time emphasizing confident predictions. This prior is theoretically motivated by the probability density function of Gaussian distribution.
In NAC, we propose three simple but effective data perturbation operations to generate native anomaly examples based on original time series sub-sequences. A new supervised training branch is devised to further calibrate the learned normality to be discriminative w.r.t. these anomaly-informed samples. 
By jointly optimizing these two components, our calibrated one-class classification model COUTA is enforced to be robust to anomaly contamination and discriminative to primeval anomalous behaviors, thus producing a more accurate data normality boundary, as depicted in Fig. \ref{fig:demonstration} (b).

As an example, we use the \textit{Omi-4} server data of the real-world Application Server Dataset (\textit{ASD}) \cite{li2021multivariate} to have a straightforward demonstration of the benefit of each calibration to our model in Fig. \ref{fig:example}, where Fig. \ref{fig:example} (a) visualizes the time series data (five representative dimensions out of the original nineteen dimensions are selected) while Fig. \ref{fig:example} (b) displays the new feature spaces learned by four different one-class classification models. 
It is clear that the canonical one-class classification (i.e., COUTA without calibration) fails to identify two anomalies (Anom1 and Anom2), and the hypersphere formed by normal data is also biased.
By contrast, adding either NAC or UMC all effectively calibrates the normality of the data, 
resulting in better discrimination of real anomalies from normal data. As a result, by adding both calibrations, COUTA can easily differentiate all three anomalies of diverse abnormal behaviors.

The contributions are summarized as follows.
\begin{itemize}
	\item We propose the calibrated one-class classification method COUTA, which calibrates one-class learning using prediction uncertainties and native anomalies. These two calibrations result in a contamination-tolerant, anomaly-informed COUTA.
	
	\item We propose the uncertainty modeling-based calibration, UMC. It restrains irregular noisy training data, while at the same time encouraging confident predictions on regular samples to ensure an effective normality learning process. This largely alleviates the negative impact of anomaly contamination.
	
	\item We propose the native anomaly-based calibration, NAC. It generates different types of anomaly examples and wields them to calibrate one-class representations for learning a more	precise abstraction of normality and a clearer normality boundary. 
	
\end{itemize}

Extensive experiments show that: (1) COUTA substantially outperforms 15 state-of-the-art competing methods on 10 real-world datasets and averagely achieves over 11\% improvement; and (2) COUTA is also with several desired properties including generalization capability in identifying different anomaly types, favorable robustness to anomaly contamination, and good scalability w.r.t. both length and dimensionality of time series.



\section{Related Work}


\subsection{Anomaly Detection in Time Series}

Time series anomaly detection is an old discipline.
There is a long list of traditional methods in the literature using various techniques like decomposition, clustering, distance, and pattern mining. 
Besides, traditional time series prediction models such as moving average, autoregressive, and their multiple variants 
are adapted to detect anomalies by comparing the predicted values and the real ones. Reviews of traditional time series anomaly detection can be found in \cite{gupta2013outlier}, and we also recommend some comprehensive time series anomaly detection benchmark studies on univariate data \cite{paparrizos2022tsb}, multivariate data \cite{schmidl2022evaluation}, and explainability \cite{jacob2021exathlon}.  


Deep learning fuels many deep time series anomaly detection methods. 
They use generative one-class learning models to restore input data or predict future data as precisely as possible. Prior work generally categorizes these current studies into reconstruction- and prediction-based methods. 
As the training set is dominated by normal data, reconstruction/prediction errors can indicate abnormal degrees. The core insight of these methods is to implicitly model normal patterns and behaviors via the rationale behind restoring or forecasting. 
The pioneer in this research line \cite{malhotra2016lstmed} uses the Long-Short Term Memory (LSTM) network in an encoder-decoder structure. 
In recent years, the data mining community has made tremendous efforts to successfully enhance the performance of this pipeline by devising various advanced network structures and new reconstruction/prediction learning objectives. 
A number of studies focus on capturing more comprehensive temporal and inter-variate dependencies by using graph neural networks \cite{deng2021gdn,zhao2020gat}, convolutional kernels \cite{zhang2019mscred,campos2021vldb}, and variational Autoencoders \cite{su2019omni,li2021multivariate}. 
Besides, adaptive memory network \cite{zhang2022adaptive} and hierarchical structure-based multi-resolution learning \cite{zhang2019mscred,huang2022efficient,shen2020thoc} are used to better handle diverse normal patterns.  
Some other methods use adversarial training in Autoencoder \cite{kieu2022robust,audibert2020usad,liu2022tkde} or generative adversarial network \cite{zhao2022crowdsourcing,liu2022tkde,du2021tkde} 
that introduce regularization into the learning process to alleviate the overfitting problem. Additionally, ensemble learning is also explored in \cite{campos2021vldb,kieu2019ensemble}. 
A very recent work \cite{tuli2022tranad} employs Transformer \cite{vaswani2017transformer} to effectively model long-term trends in time series data, and several tools are used as building blocks to further enhance the detection model, including adversarial training to amplify reconstruction errors, self-conditioning for better training stability and generalization, model agnostic meta-learning to handle the circumstance that only limited data are available. 
It is noteworthy that some studies propose new anomaly scoring strategies to replace reconstruction or prediction errors. A Transformer-based method \cite{xu2022transformer} uses association variance as a novel criterion to measure abnormality, and the literature \cite{feng2021time} employs the Bayesian filtering algorithm for anomaly scoring. 
Various advanced techniques are equipped in this pipeline, achieving state-of-the-art performance. Nevertheless, these methods may still considerably suffer from the presence of anomaly contamination and the absence of knowledge about anomalies. We below review techniques related to solving these two key problems.



\subsection{Anomaly Contamination and Label-noise Learning}

A few anomaly detection methods consider the anomaly contamination problem. The literature \cite{du2021tkde,pang2018learning,pang2020self} filters possible anomalous samples via self-training. An additional Autoencoder is used in \cite{kieu2022robust} to obtain a clean set of time series data before the training process. 
A recent work \cite{qiu2022latent} jointly infers binary labels to each sample (normal vs. anomalous) while updating the model parameters. This work applies two coupled losses that are respectively designed for normal and anomalous data.
These methods attempt to use the abnormality derived from the original/additional one-class learning component to filter these noises. However, these filtering processes also suffer from the anomaly contamination problem, and these methods may also wrongly filter some hard normal samples which are important to the network training. 
This problem is also related to label-noise learning or inaccurate supervision because these hidden anomalies are essentially training data with wrong labels. This topic is under the big umbrella of the weakly-supervised paradigm. A survey \cite{han2020survey} categorizes the methodology of this research line into three perspectives, i.e., data, learning objective, and optimization policy. 
The proposed uncertainty modeling-based calibration in the one-class learning objective also contributes a novel solution to this research line. 


\subsection{Self-supervised Anomaly Detection}
Creating supervisory signals from the data itself is an essential idea in self-supervised learning. 
Inspired by a number of self-supervised methods in image anomaly detection \cite{chai2022improving}, some recent studies are designed for time series data. They assign class labels to different augmentation operations (e.g., adding noise, reversing, scaling, and smoothing) \cite{zhang2022adaptive}, neural transformations \cite{qiu2021neural}, contiguous and separate time segments \cite{deldari2021time}, or different time resolutions \cite{huang2022efficient}. 
However, although these proxy tasks can help to better learn data characteristics, these tasks still fail to provide information related to anomalies. They may neglect that it is also possible to reliably simulate abnormal behaviors in time series via simple data perturbation. We explore this direction in our proposed method, showing that these dummy anomaly examples can greatly benefit the learning process.  


\subsection{Anomaly Exposure}
Providing extra anomaly information is a direct solution to address the absence of knowledge about anomalies. 
This idea is initially proposed by a work named outlier/anomaly exposure \cite{hendrycks2018oe}. This study employs data from an auxiliary natural dataset as manually introduced out-of-distribution examples. 
Our work is fundamentally different from \cite{hendrycks2018oe}. We create dummy anomaly examples by performing data perturbation on original data instead of taking data samples from a supplementary nature dataset. A concurrent study \cite{cai2022perturbation} also works on perturbation learning for anomaly detection in images, which constructs a perturbator and a classifier to perturb data with minimum efforts and correctly classify the normal data and perturbed data into two classes. 
Besides, DROCC \cite{goyal2020drocc} also adaptively generates anomalous points while training via a gradient ascent phase reminiscent of adversarial training.
Different from these studies, we propose tailored perturbation methods for time-series data by fully considering the unique characteristics and clear definitions of anomalies in time series.

\section{The Proposed Method: COUTA}

\subsection{Problem Formulation}

Let $\mathcal{X} = \langle \bm{x}_1, \bm{x}_2, \cdots, \bm{x}_N \rangle$ be a time series dataset, and $\mathcal{X}$ is an ordered sequence of $N$ observations.
Each observation in $\mathcal{X}$ is a vector described by $D$ dimensions (i.e., $\bm{x}_t \in \mathbb{R}^D, \forall \bm{x}_t \in \mathcal{X}$). Dataset $\mathcal{X}$ is termed as multivariate time series when $D>1$, and the dataset is reduced to the univariate setting if $D=1$.
Unsupervised time series anomaly detection $f$ is to measure the abnormal degree of each observation and give anomaly scores without accessing any label information, i.e., $f: \mathcal{X} \mapsto \mathbb{R}$. Higher anomaly scores indicate a higher likelihood to be anomalies.

We consider a local contextual window of each observation to model their temporal dependence. Specifically, a sliding window with length $l$ and stride $r$ is used to transform the training set into a collection of sub-sequences $\mathcal{S} = \{ \bm{s}_1, \bm{s}_{1+r}, \cdots \}$, where $\bm{s}_t = \langle \bm{x}_t, \bm{x}_{t+1}, \cdots, \bm{x}_{t+l-1} \rangle$. This is a common practice in most deep time-series anomaly detection methods \cite{tuli2022tranad,campos2021vldb,su2019omni,li2021multivariate}. During the inference stage, the testing set is also split into sub-sequences using the same window length $l$, and the sliding stride is set to 1. The anomaly detection model evaluates the abnormal degree of each sub-sequence, and the anomaly score is assigned to the last timestamp of each sub-sequence. We use 0 to pad the beginning $l-1$ timestamps to obtain the final anomaly score list.

        
			


\subsection{Overall Framework}

\begin{figure*}[t]
	\centering
	\includegraphics[width=0.9\textwidth]{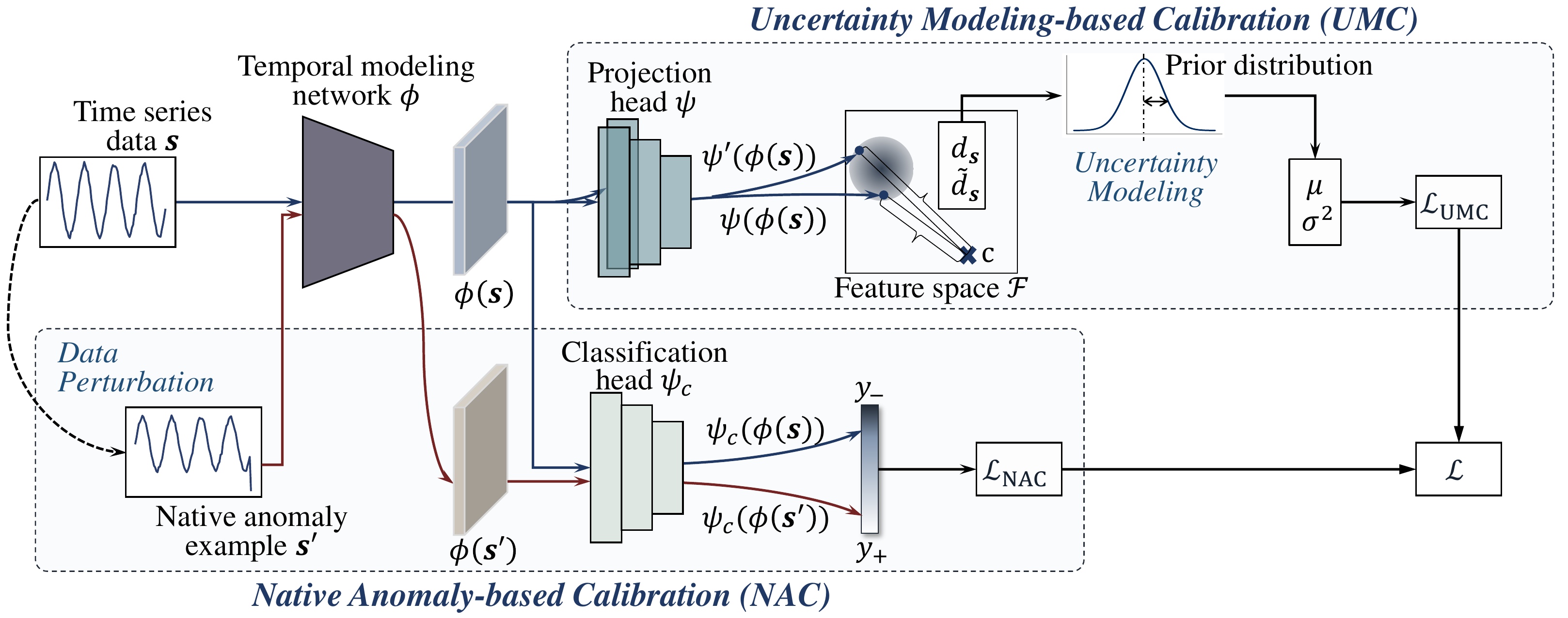}
	\caption{The framework of COUTA. 
	}
	\label{fig:framework}
\end{figure*}

The overall framework COUTA is shown in Fig. \ref{fig:framework}.
A temporal modeling network $\phi$ is used to model time-axis dependency and inter-variate interactions.
We opt for Temporal Convolutional Network (TCN) \cite{bai2018tcn} as the temporal modeling network $\phi$ in COUTA. TCN is more time-efficient than traditional RNN-based structures, and it can better capture local time dependency due to the usage of convolutional kernels.
We further use a lightweight projection head $\psi$ to map data into an $H$-dimensional feature space $\mathcal{F} \subset \mathbb{R}^H$. 
The multi-layer perceptron network structure is employed in $\psi$. The whole representation process can be denoted as $\psi(\phi(\cdot))$. 
We aim to map the training sample $\bm{s}$ into a compact hypersphere with center $\bm{c}$ upon the feature space $\mathcal{F}$. We devise the Uncertainty Modeling-based Calibration (UMC) in COUTA. 
Specifically, a prior probability distribution is imposed on the one-class distance. We set a bypass in $\psi$ by using an additional layer, producing an extra prediction $\psi'(\phi(\bm{s}))$. Two distance values (i.e., $d$ and $\tilde{d}$) are used to obtain the mean and the variance of the prior distribution. 
Our one-class learning objective $\mathcal{L}_{\text{UMC}}$ is calibrated to adaptively penalize uncertain predictions and simultaneously encourage confident predictions, thus accomplishing the masking of anomaly contamination in the training set.
Besides, we propose Native Anomaly-based Calibration (NAC) to introduce knowledge about time series anomalies to the one-class learning process. 
Native anomaly example is created based on original time series data via tailored data perturbation operations. A new supervised training branch with a classification head $\psi_c$ is added to empower COUTA to discriminate abnormal behaviors in time series via loss function $\mathcal{L}_{\text{NAC}}$. We also employ the multi-layer perceptron structure in $\psi_c$. 

The final loss function is computed by assembling: 
\begin{equation}\label{eqn:loss0}
	\mathcal{L} = \mathcal{L}_{\text{UMC}} + \alpha \mathcal{L}_{\text{NAC}},
\end{equation}
where $\alpha$ is a parameter to adjust the weight of the native anomaly-based classification branch. 


The learnable parameters within the neural network are jointly trained by loss function $\mathcal{L}$. During the inference stage, the testing set is also pre-processed to sub-sequences, and COUTA measures abnormal degrees of input sub-sequences according to their deviation from the learned normality model (i.e., the hypersphere). 


\subsection{Calibrated One-class Classification}

\subsubsection{UMC for Contamination-tolerant One-class Learning }

COUTA aims at learning a hypersphere with the minimum radius that can well enclose the training data upon the feature space $\mathcal{F}$. Therefore, the data normality can be explicitly defined as this hypersphere, and the distance to the hypersphere center can faithfully indicate the degree of data abnormality. 
This basic goal is the same as SVDD \cite{tax2004svdd} (a popular technique in one-class classification). The traditional SVDD algorithm relies on the kernel trick. 
As has been done in a recent extension \cite{ruff2018deepsvdd}, after mapping original data to a new feature space via neural network $\phi$ and $\psi$, the canonical one-class loss function can be defined as 
\begin{equation}\label{eqn:svdd}
	\mathcal{L}_{\text{canonical}} = 
	\mathbb{E}_{\bm{s} \sim \mathcal{S}} \Big [ \big\Vert  \psi \big( \phi(\bm{s}) \big) - \bm{c} \big\Vert^2 \Big ],
\end{equation}
where $\bm{c}\in\mathbb{R}^H$ is the hypersphere center upon the feature space $\mathcal{F}$.


Anomaly contamination of the training set is essentially noisy data that have very rare abnormal behaviors and inconsistent patterns, and thus the one-class classification model tends to output unconfident predictions on these noisy data. 
Therefore, to address the anomaly contamination problem, we can give a relatively mild penalty to predictions that are with high model uncertainty, thus masking anomaly contamination in a soft manner. On the other hand, to ensure effective optimization of hard normal samples, the one-class classification model should also be encouraged to output confident predictions. 
The learning objective in Equation (\ref{eqn:svdd}) is basically defined according to the distance to the hypersphere center.
Therefore, the core idea is to impose a prior (i.e., Gaussian) distribution $\mathcal{N}_{\bm{s}}(\mu, \sigma^2)$ to the single distance value $d_{\bm{s}} = \Vert \psi(\phi(\bm{s})) - \bm{c} \Vert^2$, and the variance $\sigma^2$ of the distribution can represent the model uncertainty. 
Hence, to fulfill uncertainty modeling-based calibration, we need to answer two questions, i.e., \textit{how to design a new learning objective that can handle distance distribution} and \textit{how to extend the single distance value to its prior distribution}.



We first consider the design of our new one-class loss function. Given a prior distribution of the one-class distance, we need to maximize the probability of the distance being zero, instead of simply minimizing a single distance value. Based on the probability density function of the Gaussian distribution, the learning objective of the distance distribution $\mathcal{N}_s(\mu, \sigma^2)$ of sub-sequence $\bm{s}$ can be defined to maximize the following function:
\begin{equation}
	J = \frac{1}{ \sqrt{2\pi \sigma^2}} e^{-\frac{1}{2}  (\frac{ \mu  }{ \sigma})^2}.
\end{equation}


We can further derive the following function:
\begin{equation}\label{eqn:objective}
	\begin{split}
		\log J = -\frac{1}{2 \sigma^2} \mu^2 - \frac{1}{2} \log 2\pi \sigma^2.
	\end{split}
\end{equation}

We omit $2\pi$ and use $\zeta$ to indicate $\log \sigma^2$, and the learning objective of the Gaussian distribution $\mathcal{N}_{\bm{s}}(\mu, \sigma^2)$ is equivalent to minimize the following loss value:
\begin{equation}\label{eqn:ell}
	\ell(\bm{s}) = \frac{1}{2} e^{-\zeta} \mu^2 
	+ \frac{1}{2} \zeta. 
\end{equation}


We then address how to extend the single output of a distance value to a Gaussian distribution. One direct solution is to employ the ensemble method to obtain a group of predictions, thus estimating the mean and the variance of the distribution. 
However, the GPU memory consumption and the computational effort might be costly when there are many ensemble members.
The essence of the Gaussian distribution is to find the variance, that is, the ensemble process does not need a heavy computational overhead. Recall that a lightweight projection head $\psi$ is used after the temporal modeling network $\phi$, we can set a bypass hidden layer in $\psi$, yielding an additional projection output (denoted by $\psi'(\phi(\bm{s}))$). We calculate one-class distance values of $\psi(\phi(\bm{s}))$ and $\psi'(\phi(\bm{s}))$ as
\begin{equation}\label{eqn:d}
	d_{\bm{s}} = \Vert \psi(\phi(\bm{s})) - \bm{c} \Vert^2, \tilde{d}_{\bm{s}} = \Vert \psi'(\phi(\bm{s})) - \bm{c} \Vert^2.
\end{equation}


We further employ $d_{\bm{s}} + \tilde{d}_{\bm{s}}$ and $(d_{\bm{s}} - \tilde{d}_{\bm{s}})^2$ to delegate $\mu^2$ and $\zeta$, respectively. 
Therefore, the loss function of one-class classification with uncertainty modeling-based calibration can be further derived as follows:
\begin{equation}\label{eqn:lossn}
	\mathcal{L}_{\text{UMC}} =  \mathbb{E}_{\bm{s} \sim \mathcal{S}} \Big[
	\frac{1}{2} e^{- ( d_{\bm{s}}- \tilde{d}_{\bm{s}} ) ^2} (d_{\bm{s}} + \tilde{d}_{\bm{s}} ) 
	+ 
	\frac{1}{2}( d_{\bm{s}} - \tilde{d}_{\bm{s}}  )^2
	\Big].
\end{equation}

\begin{figure}[t]
	\centering
	\includegraphics[width=0.3\textwidth]{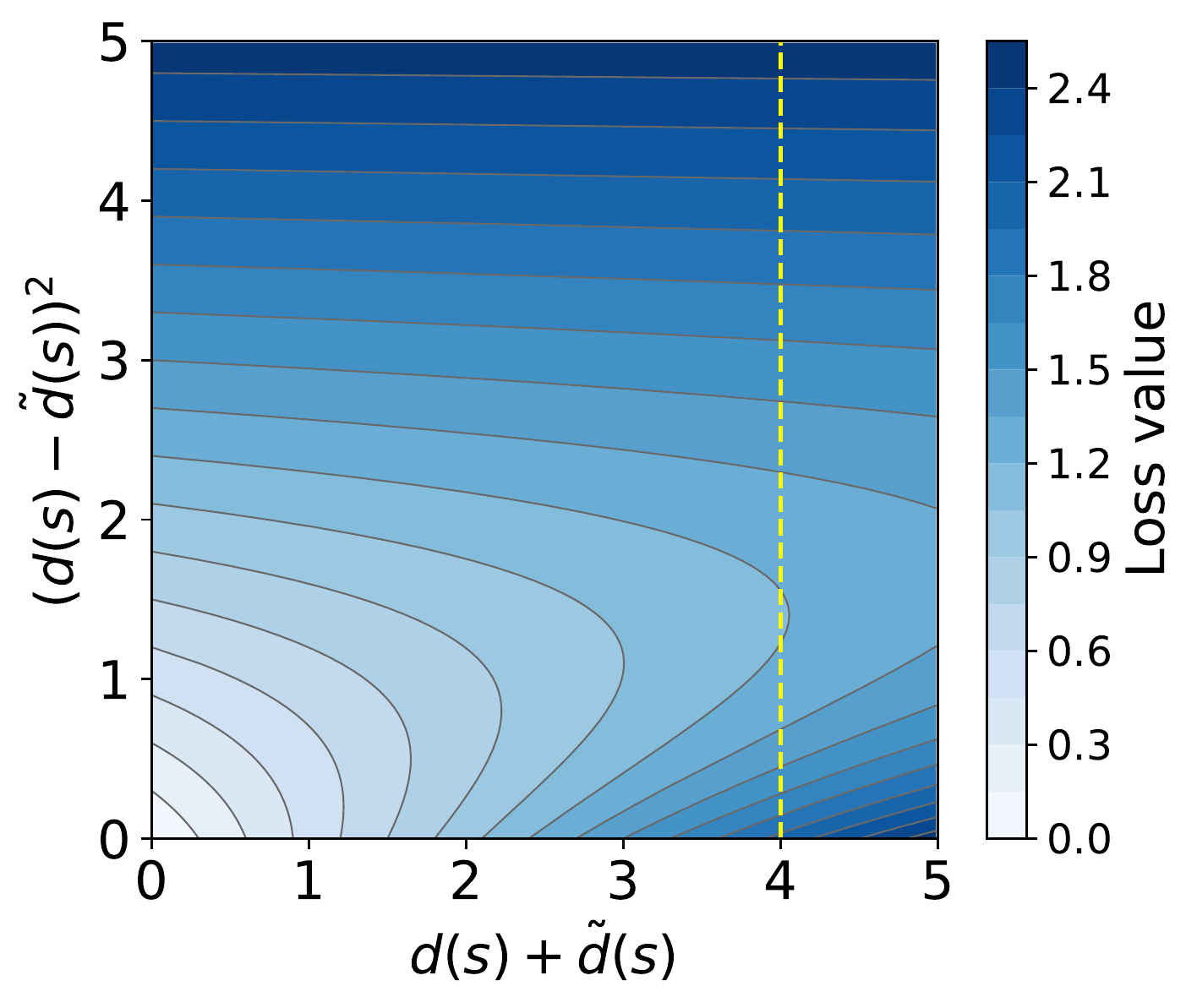}	\caption{Loss values in $\mathcal{L}_{\text{UMC}}$ w.r.t. $d_{\bm{s}} + \tilde{d}_{\bm{s}}$ and $(d_{\bm{s}} - \tilde{d}_{\bm{s}})^2$. As indicated by the yellow line, the penalty of a fixed $d_{\bm{s}} + \tilde{d}_{\bm{s}}$ is first adjusted to more mild levels with the increase of the uncertainty $(d_{\bm{s}} - \tilde{d}_{\bm{s}})^2$, while the loss function further penalizes heavily when the uncertainty reaches a high value.}
	\label{fig:loss}
\end{figure}

Our loss function in Equation (\ref{eqn:lossn}) can mask anomaly contamination and weaken their contribution by assigning mild penalties. The one-class classification model tends to output inconsistent predictions on these hidden anomalies, i.e., $(d_{\bm{s}}- \tilde{d}_{\bm{s}})^2$ is high. 
Therefore, the loss value $d_{\bm{s}} + \tilde{d}_{\bm{s}}$ is adjusted to a lower level by its coefficient $e^{- (d_{\bm{s}}- \tilde{d}_{\bm{s}})^2 }$. 
On the other hand, the second term also penalizes high uncertainty, which encourages more confident predictions to ensure the effective optimization of hard samples. 
Fig. \ref{fig:loss} visualizes $\mathcal{L}_{\text{UMC}}$ of Equation (\ref{eqn:lossn}) by presenting how loss values change w.r.t. $d_{\bm{s}}+\tilde{d}_{\bm{s}}$ and $(d_{\bm{s}} - \tilde{d}_{\bm{s}})^2$. As expected, this loss function adaptively adjusts the loss value of the data sample with high uncertainty and concurrently impels more confident predictions.

Note that our method is fundamentally different from the existing work based on Gaussian processes \cite{kemmler2013one}. This work transfers the one-class classification problem to Gaussian Process Regression and approximates Gaussian Process classification with Laplace approximation or expectation propagation. Differently, our work imposes a Gaussian distribution to the one-class distance value, thus modeling the uncertainty during the one-class learning process.

\subsubsection{NAC for Anomaly-informed One-class Learning}\label{sec:nac}

To introduce knowledge about anomalies, we propose to offer dummy anomaly examples to the one-class classification model. 
We introduce several tailored perturbation operations to generate native anomaly examples based on original time series data.


\begin{figure}[t]
	\centering
	\includegraphics[width=0.49\textwidth]{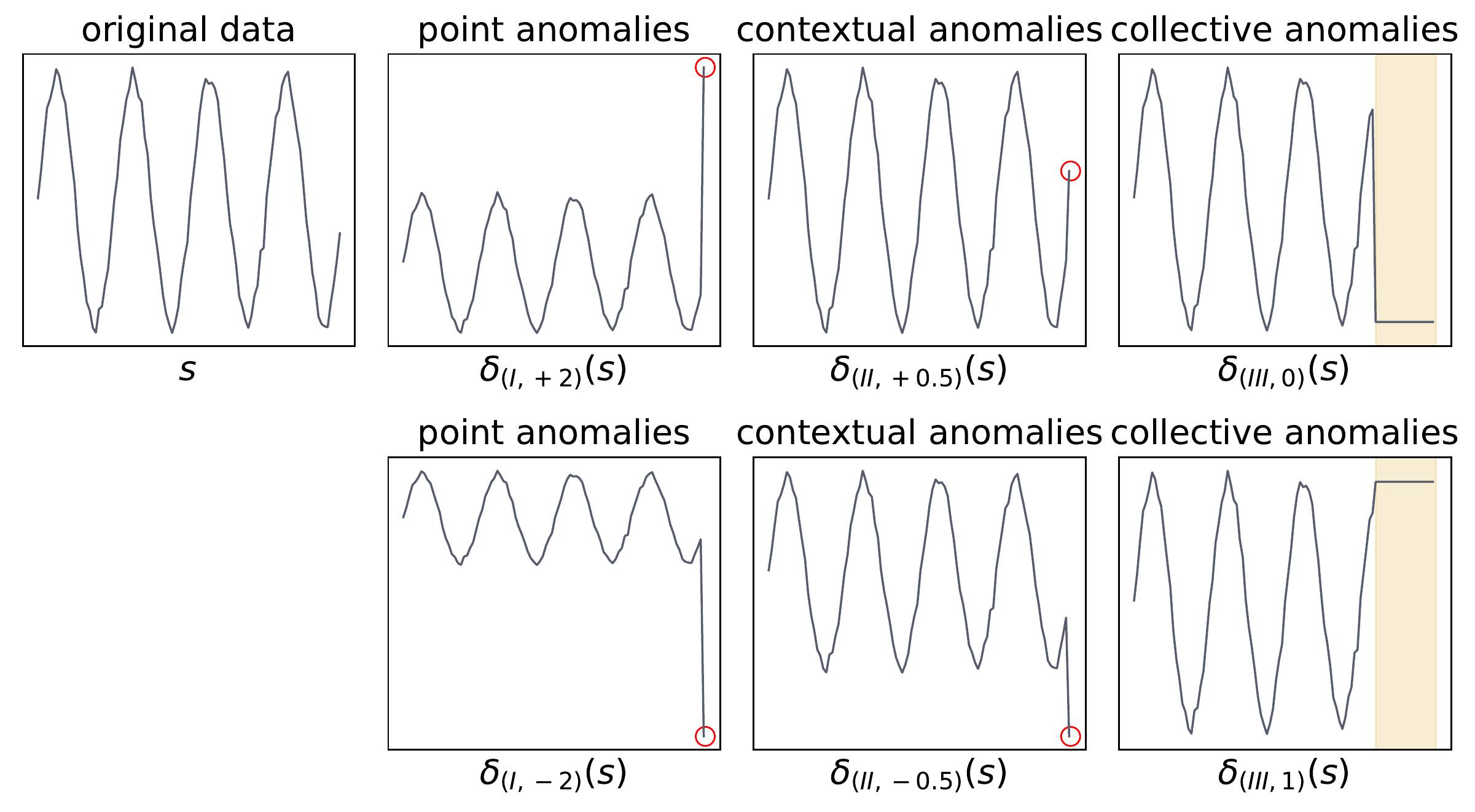}
	\caption{Native anomaly examples generated from a time series sub-sequence $\bm{s}$ by six perturbation functions in $\Omega$. }
	\label{fig:anom_example}
\end{figure}

Data perturbation $\delta$ modifies the input time series sub-sequence $\bm{s}$ via a specific operation to obtain a new sub-sequence $\bm{s}'$, such that $\bm{s}'$ contains realistic abnormal behaviors of time series. $\bm{s}$ and $\bm{s}'$ are with the same shape. 
According to the definitions and characterizations of three basic time series anomaly types (i.e., point, contextual, and collective anomalies), we define the following data perturbation operations. 
\begin{itemize}
	\item $\delta_{(\text{I}, \gamma)}(\bm{s})$: Data perturbation is performed on the last observation of the input sub-sequence $\bm{s}$. Data values of a group of randomly selected dimensions of the last observation are replaced with a global extreme value $\gamma$. A new sub-sequence with a point anomaly can be created.
	\item $\delta_{(\text{II}, \gamma)}(\bm{s})$: Data perturbation is also applied to randomly selected dimensions of the last observation of the input sub-sequence $\bm{s}$. We use an offset $\gamma$ based on the mean of the previous $k$ values to pad these selected places. This perturbation method produces contextual anomalies. We use $k=10$ by default.
	\item $\delta_{(\text{III}, \gamma)}(\bm{s})$:  We also randomly choose a group of dimensions, but this perturbation operation acts on a segment of the input sub-sequence $\bm{s}$.
	The right side is fixed as the last observation, and the segment length is randomly sampled from the range $[1, w]$. These values are replaced with an outlier value $\gamma$. This perturbation process produces collective anomalies. We use $w=0.5l$ by default.
\end{itemize}

In practice, each of the above operations is deployed with two $\gamma$ values to simulate anomalies at two extremes. As time series data is first preprocessed via data normalization to $[0, 1]$, we use $\gamma=+2,-2$ in $\delta_{\text{I}}$, $\gamma=+0.5, -0.5$ in $\delta_{\text{II}}$, $\gamma=1, 0$ in $\delta_{\text{III}}$. We define a pool $\Omega$ containing these six data perturbation functions, i.e.,
\begin{equation}
	\begin{split}
		\Omega = \big\{ 
		\delta_{(\text{I}, +2)}, 
		\delta_{(\text{I}, -2)}, 
		\delta_{(\text{II}, +0.5)}, 
		\delta_{(\text{II}, -0.5)}, 
		\delta_{(\text{III}, 0)}, \delta_{(\text{III}, 1)}
		\big \}	.
	\end{split}
\end{equation}
These six perturbation functions can simulate abnormal behaviors in time series data. Fig. \ref{fig:anom_example} presents a base time series sub-sequence $\bm{s}$ and corresponding native anomaly examples generated via these six data perturbation operations within $\Omega$.

%

A new set of dummy anomaly examples $\mathcal{S}'$ with size $\beta |\mathcal{S}|$ is generated, which is denoted as follows.

\begin{equation}
	\mathcal{S}' = \big \{ \delta(\bm{s}) | \bm{s} \sim \mathcal{S}, \delta \sim \Omega \big \},
\end{equation}
where original sub-sequence $\bm{s}$ and function $\delta$ are randomly sampled from the training set $\mathcal{S}$ and the operation pool $\Omega$.

We then set a new supervised training branch to calibrate COUTA such that our one-class classifier can discriminate primitive anomalies in $\mathcal{S}'$. 
We use an extra lightweight classification head $\psi_c$ following the temporal modeling network $\phi$ to transfer each data sample to a score. $\psi_c$ also uses the multi-layer perceptron structure.
Mean squared error is employed to regress the output of sub-sequences of the original set to $y_{-}$ and the output of those anomaly examples to $y_+$. The loss function of this branch is defined as:
\begin{equation}\label{eqn:lossc}
\begin{split}
	\mathcal{L}_{\text{NAC}} = \mathbb{E}_{\bm{s} \sim \mathcal{S}  \cup \mathcal{S}'}  
	\Big [
	&\mathbbm{1}_{\bm{s} \in \mathcal{S}} \big( \psi_c(\phi(\bm{s})) - y_- \big)^2 + \\
	&\mathbbm{1}_{\bm{s} \in \mathcal{S}'} \big( 
	\psi'(\phi(\bm{s}))  - y_+
	\big)^2
	\Big ],
\end{split}
\end{equation}
where $y_{+}=1$ and $y_{-}=-1$ are used in our implementation.

Although these manually defined perturbation operations seem to be old-school compared to deep generation methods, this way is simple enough to generate dummy anomaly examples with abnormal behaviors of time series. 
Exposing these native anomaly examples to the one-class classification model via this training branch can lead to a more precise abstraction of data normality and a clearer boundary of the normality hypersphere.

It is noteworthy that COUTA also contributes to better detection performance when there is no overlap between these simulated anomalies and the specific anomaly type in target datasets. 
NAC can be seen as a self-supervised learning process, which can effectively strengthen one-class learning. 
Self-supervised learning defines various transformations and designs pretext tasks to classify or compare these transformations. 
Similarly, COUTA employs tailored anomaly-aware transformations for time series data and trains neural networks to discriminate transformed data from original data. By harnessing pretext tasks, self-supervised learning can embed data semantics into representations. Likewise, COUTA also better learns temporal patterns and inter-variant dependency within input time-series data, which poses a positive effect on the one-class learning process. Besides, we define tailored anomaly-aware transformations for time series data, i.e., NAC can achieve anomaly-informed learning that better suits the downstream anomaly detection task.


\subsection{Anomaly Scoring}

The learned hypersphere upon the feature space $\mathcal{F}$ can explicitly represent the data normality, and data abnormality can be simply defined as the Euclidean distance to the hypersphere center $\bm{c}$. As each distance value is extended to a Gaussian distribution to express model uncertainty in our calibrated one-class classification model, we employ two distance values to define the abnormal degree. Given the optimized network including the temporal modeling network $\phi^{*}$ and the projection heads $\psi^{*}$ and $\psi'^{*}$, the anomaly score of a sub-sequence $\bm{s}$ is calculated as follows. 
\begin{equation}
	\begin{split}
		f(\bm{s}) &= d_{\bm{s}} + \tilde{d}_{\bm{s}} \\
		& = \Vert \psi^{*}(\phi^{*}(\bm{s})) - \bm{c} \Vert + \Vert \psi'^{*}(\phi^{*}(\bm{s})) - \bm{c} \Vert.
	\end{split}
\end{equation}


\subsection{Discussion} 

This section further discusses our considerations and implementation details.

\textbf{Discriminative vs. Generative.}
COUTA is in a discriminative manner instead of using mainstream reconstruction- or prediction-based generative models. Compared to the autoencoder structure that is composed of an encoder phase and a decoder phase, COUTA does not need to reconstruct the encoded feature back to the original shape, which is more time efficient. Moreover, COUTA learns a compact hypersphere, which is an explicit way to model data normality. That is, the optimization is directly related to anomaly detection rather than implicitly behind the rationale of data reconstruction or forecasting.

\textbf{The Choice of Hypersphere Center.}
Arguably, including $\bm{c}$ as an optimization variable will lead to a trivial solution, i.e., all learnable parameters in network $\phi$ and $\psi$ are zero \cite{ruff2018deepsvdd}. Hence, following \cite{ruff2018deepsvdd}, we use the initialized network to perform a forward pass on all training data, and $\bm{c}$ is set as the averaged representation, i.e.,
$\bm{c} = \frac{1}{|\mathcal{S}|} \sum_{\bm{s} \in \mathcal{S}} \psi_0(\phi_0(\bm{s}))$, 
where $\phi_0$ and $\psi_0$ are initialized neural networks before gradient optimization. It is an empirically good strategy, which makes the optimization process converge quickly and also avoids the above ``hypersphere collapse'' problem.

\textbf{Anomaly Types in Native Anomaly Generation.}
We define six perturbation functions with fixed parameters in $\Omega$. This component is also a good handle to embed readily accessible prior knowledge into the learning process. Some specific real-world applications may have their own definitions of anomalies.
For example, IT operations in data centers focus on collective (pattern) anomalies and often omit point anomalies. 
These collective anomalies may indicate real severe faults, possible downtime of running services, and unreasonable increase of system overhead, but point anomalies are often noises induced by many possible factors of system instability. Thus, data perturbation can be designed to generate corresponding anomalies of real interests. 
Note that it is also a limitation of this calibration method. These data perturbation methods are designed based on general anomaly definitions. They may bring negative effects if these generated dummy anomalies are essentially normal in target systems. Nevertheless, we empirically prove that these data perturbation methods are effective in the vast majority of real-world datasets from different domains.


\textbf{Anomaly Scoring Strategy.}
The anomaly scoring function does not use the prediction results reported by the classification head $\psi_c$. This branch is used to assist the optimization of the temporal modeling network $\phi$ by providing knowledge about anomalies. We simply treat all training data as a normal class in this branch, which means this learning task might also be misled by the anomaly contamination problem. Besides, these dummy anomaly examples might not always be reliable due to the limitation analyzed above. Simply adding the output of this branch may downgrade the detection performance, and it is also challenging to devise a good ensemble method to integrate the outputs of two branches. Therefore, we leverage the one-class learning results calibrated by both UMC and NAC, i.e., the distance to the hypersphere center, to measure data abnormality in our anomaly scoring function.

\section{Experiments}\label{sec:exp}

In this section, we first introduce the experimental setup, and then we conduct experiments to answer the following questions. 

\begin{itemize}
	\item \textbf{Effectiveness:} How accurate are the anomaly detection results computed by COUTA and current state-of-the-art methods on real-world datasets?
	\item \textbf{Generalization ability:} Can COUTA generalize to different types of time series anomalies?
	\item \textbf{Robustness:} How does the robustness of COUTA w.r.t. various anomaly contamination levels of the training set?
	\item \textbf{Scalability:} Is COUTA more time-efficient compared to existing methods?
	\item \textbf{Sensitivity:} How do hyper-parameters of COUTA influence the detection performance?
	\item \textbf{Ablation study:} Do the proposed calibration methods contribute to sufficiently better detection performance?
\end{itemize}

\subsection{Experimental Setup}

\subsubsection{Datasets and Preprocessing Methods}

\begin{figure}[t]
	\centering
	\includegraphics[width=0.49\textwidth]{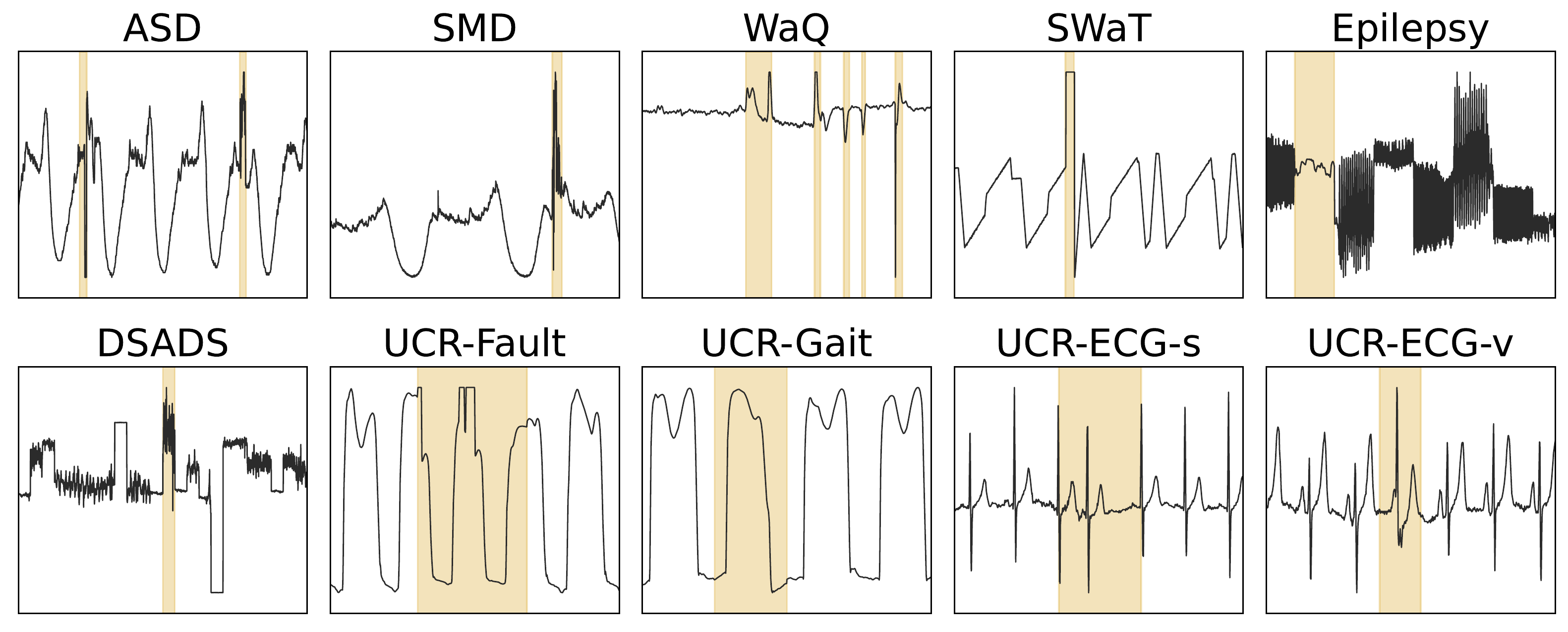}
	\caption{Anomaly segments in ten real-world datasets.  
	}
	\label{fig:anomalies}
\end{figure}

Ten publicly available datasets are used in our experiments, including six multivariate datasets and four univariate datasets. 
Application Server Dataset (\textit{ASD}) and Server Machine Dataset (\textit{SMD}) are data of IT operations, in which each variate denotes the status and resource utilization of servers in a cluster. Secure Water Treatment dataset (\textit{SWaT}) and Water Quality dataset (\textit{WaQ}) are industrial data. The dimensions of \textit{SWaT} are different sensors and actuators, and \textit{WaQ} is to identify undesirable variations in the water according to a group of chemistry and physical metrics. 
Epilepsy seizure dataset (\textit{Epilepsy}) and Daily and Sports Activities Dataset (\textit{DSADS}) are motion sensor data. \textit{Epilepsy} is to detect epilepsy seizures from three activities including walking, running, and sawing. Following \cite{zhang2022adaptive}, anomalies in \textit{DSADS} are rare and intense activities, while the remaining activities are defined as normal data.
\textit{UCR-Fault} and \textit{UCR-Gait} come from a biomechanics lab, recording the gait of someone walking on a force plate. Anomalies in \textit{UCR-Fault} are faults in the left foot sensor, and anomalies in \textit{UCR-Gait} are the gait of the individual who has Huntington’s disease.
\textit{UCR-ECG-s} and \textit{UCR-ECG-v} are ECG data, respectively containing supraventricular beats and ventricular beats. 
We report their statistics in Table \ref{tab:dataset}, and Fig. \ref{fig:anomalies} shows representative anomaly segments of each dataset.

These datasets are selected due to the following considerations: (i) These data are from various real-world applications; (ii) 
They are of different lengths, dimensionality, and anomaly ratios; 
(iii) Anomalies in these datasets can be categorized into different types;   
(iv) These datasets are frequently used in the literature. 
\textit{ASD}, \textit{SMD} and \textit{SWaT} are benchmark datasets in this research line \cite{su2019omni,li2021multivariate,tuli2022tranad,campos2021vldb,liu2022tkde,deng2021gdn}. 
\textit{WaQ} is included in a benchmark paper \cite{lai2021revisiting}. \textit{Epilepsy} and \textit{DSADS} are datasets taken from the time series classification task by treating semantically abnormal class(es) as anomalies, as has been done in many prior anomaly detection studies \cite{qiu2021neural,zhang2022adaptive,xu2023icml,xu2023rosas}. 
\textit{UCR} datasets are from the latest time series anomaly archive \cite{wu2021current}.

\textit{ASD}, \textit{SMD}, \textit{SWaT}, \textit{WaQ}, and \textit{UCR} datasets have pre-defined training-testing split. In terms of \textit{Epilepsy} and \textit{DSADS}, we use 60\% data as the training set and the remaining 40\% as the testing set while maintaining the original anomaly proportion.
Following \cite{garg2021evaluation}, the minimum and maximum values per dimension of the training set are obtained to conduct min-max normalization, and the data values in the testing set are clipped to $[\min-4, \max+4]$ to prevent excessively large values skewing anomaly scoring process. 
Deep anomaly detectors generally use a sliding window to divide time series datasets into small sub-sequences.
A sliding window with a fixed size of 100 is used for \textit{ASD}, \textit{SMD}, \textit{SWaT}, \textit{WaQ}, and four \textit{UCR} datasets. We take 1 as the sliding stride for \textit{ASD}, \textit{SMD} and respectively use 100 and 5 for large-scale datasets \textit{SWaT} and \textit{WaQ}. As for the remaining two datasets, \textit{Epilepsy} and \textit{DSADS}, the original data format is collections of divided time sequences, and thus sliding window is not required.


\begin{table}[t]
	\centering
	\caption{Dataset Information. $N$ and $D$ are length and dimensionality, \#anom denotes the number of anomalies, and \#entity is the number of sub-datasets in each data repository (each sub-dataset is trained and tested independently). 
	We report the maximum and minimum values of entities in \textit{ASD}.
	Each entity in \textit{DSADS} has the same statistics.
	The information of each entity in \textit{SMD} and four \textit{UCR} datasets are respectively reported.
  }
	\scalebox{0.88}{
	\begin{tabular}{lllll}
	\hline
	\textbf{Dataset}  & \boldmath$N$\unboldmath  & \boldmath$D$\unboldmath   
	& \textbf{\#anom}     & \textbf{\#entity} \\
	\hline
    ASD   & 11,611 - 12,960 & 19    & 55 - 441 & 12 \\
	SMD   & 57,426/57,391/57,426 & 38    & 308/198/303 & 3 \\
	SWAT  & 925,119 & 51    & 54584 & 1 \\
	WaQ   & 253,607 & 9     & 2329  & 1 \\
	DSADS & 142,500 & 45    & 9000  & 8 \\
	Epilepsy & 56,650 & 3     & 5768  & 1 \\
	\hline
	UCR-Fault & 64,000/64,000/64,000 & 1     & 111/651/801 & 3 \\
	UCR-Gait  & 65,000/65,000/65,000 & 1     & 301/301/601 & 3 \\
	UCR-ECG-s & 60,000/57,001 & 1     & 371/371 & 2 \\
	UCR-ECG-v & 40,000/40,000/80,000 & 1     & 101/101/251 & 3 \\
	\hline
	\end{tabular}}%
	\label{tab:dataset}%
\end{table}%

\subsubsection{Competing Methods}

COUTA is compared with sixteen anomaly detection methods including ARMA \cite{brockwell1991time}, OCSVM \cite{manevitz2001one},  GOAD \cite{bergman2020goad}, ECOD \cite{li2022ecod}, DAGMM \cite{zong2018deep}, 
LSTM-ED \cite{malhotra2016lstmed}, LSTM-Pr \cite{hundman2018detecting}, Tcn-ED \cite{garg2021evaluation}, DSVDD \cite{ruff2018deepsvdd}, MSCRED \cite{zhang2019mscred}, Omni \cite{su2019omni}, 
USAD \cite{audibert2020usad}, GDN \cite{deng2021gdn}, NeuTraL \cite{qiu2021neural}, TranAD \cite{tuli2022tranad}, and AnomTran \cite{xu2022transformer}.  
These competing methods include both traditional and deep approaches. Also, these competing methods employ different learning strategies (prediction, reconstruction, and discriminative one-class/self-supervised learning) and various network structures (MLP, LSTM, GRU, TCN, Transformer, convolutional net, and graph neural network). The above competitor list can well represent the state-of-the-art performance of the research line of time series anomaly detection.

\begin{table*}[htbp]
	\centering
	\caption{$F_1$ score and AUC-PR $\pm$ standard deviation of COUTA and its competitors on six multivariate datasets.}
	\scalebox{0.85}{
		\begin{tabular}
        {p{1.25cm}			| 
            p{1.18cm}p{1.18cm}p{1.18cm}
            p{1.18cm}p{1.18cm}p{1.27cm}  |
            p{1.18cm}p{1.18cm}p{1.18cm}
            p{1.18cm}p{1.18cm}p{1.27cm}
        }
        \hline
		\multicolumn{1}{c|}{\multirow{2}[0]{*}{\textbf{Methods}}} & 
		\multicolumn{6}{c|}{\boldmath$F_1$\unboldmath} & 
		\multicolumn{6}{c}{\textbf{AUC-PR}}   \\
	\cline{2-7} \cline{8-13}
		& \textbf{ASD} & \textbf{SMD} & \textbf{SWaT} & \textbf{WaQ} & \textbf{DSADS} & \textbf{Epilepsy} & \textbf{ASD} & \textbf{SMD} & \textbf{SWaT} & \textbf{WaQ} & \textbf{DSADS} & \textbf{Epilepsy} \\	
		\hline	
    ARMA & 0.474$_{\pm 0.000}$ & N/A   & N/A   & 0.647$_{\pm 0.000}$ & N/A   & 0.826$_{\pm 0.000}$ & 0.395$_{\pm 0.000}$ & N/A   & N/A   & 0.633$_{\pm 0.000}$ & N/A   & 0.801$_{\pm 0.000}$ \\
    OCSVM & 0.625$_{\pm 0.000}$ & 0.761$_{\pm 0.000}$ & 0.839$_{\pm 0.000}$ & 0.732$_{\pm 0.000}$ & 0.807$_{\pm 0.000}$ & 0.781$_{\pm 0.000}$ & 0.540$_{\pm 0.000}$ & 0.664$_{\pm 0.000}$ & 0.804$_{\pm 0.000}$ & 0.666$_{\pm 0.000}$ & 0.753$_{\pm 0.000}$ & 0.708$_{\pm 0.000}$ \\
    GOAD  & 0.827$_{\pm 0.025}$ & 0.975$_{\pm 0.019}$ & 0.831$_{\pm 0.003}$ & 0.739$_{\pm 0.052}$ & 0.723$_{\pm 0.014}$ & 0.554$_{\pm 0.147}$ & 0.834$_{\pm 0.023}$ & 0.981$_{\pm 0.013}$ & 0.799$_{\pm 0.005}$ & 0.685$_{\pm 0.034}$ & 0.676$_{\pm 0.015}$ & 0.530$_{\pm 0.128}$ \\
    ECOD  & 0.589$_{\pm 0.000}$ & 0.755$_{\pm 0.000}$ & 0.849$_{\pm 0.000}$ & 0.676$_{\pm 0.000}$ & 0.923$_{\pm 0.000}$ & 0.785$_{\pm 0.000}$ & 0.527$_{\pm 0.000}$ & 0.724$_{\pm 0.000}$ & 0.899$_{\pm 0.000}$ & 0.716$_{\pm 0.000}$ & 0.941$_{\pm 0.000}$ & 0.763$_{\pm 0.000}$ \\
    DAGMM & 0.743$_{\pm 0.105}$ & 0.652$_{\pm 0.168}$ & 0.832$_{\pm 0.010}$ & 0.592$_{\pm 0.092}$ & 0.775$_{\pm 0.067}$ & 0.793$_{\pm 0.024}$ & 0.708$_{\pm 0.117}$ & 0.581$_{\pm 0.221}$ & 0.806$_{\pm 0.011}$ & 0.539$_{\pm 0.101}$ & 0.764$_{\pm 0.091}$ & 0.774$_{\pm 0.040}$ \\
    LSTM-Pr & 0.593$_{\pm 0.011}$ & 0.860$_{\pm 0.044}$ & 0.834$_{\pm 0.055}$ & 0.532$_{\pm 0.035}$ & 0.856$_{\pm 0.011}$ & 0.661$_{\pm 0.006}$ & 0.513$_{\pm 0.013}$ & 0.858$_{\pm 0.051}$ & 0.823$_{\pm 0.101}$ & 0.463$_{\pm 0.049}$ & 0.934$_{\pm 0.010}$ & 0.513$_{\pm 0.011}$ \\
    LSTM-ED & 0.807$_{\pm 0.013}$ & 0.960$_{\pm 0.000}$ & 0.847$_{\pm 0.007}$ & 0.759$_{\pm 0.006}$ & 0.865$_{\pm 0.011}$ & 0.802$_{\pm 0.012}$ & 0.767$_{\pm 0.016}$ & 0.955$_{\pm 0.009}$ & 0.848$_{\pm 0.005}$ & 0.714$_{\pm 0.018}$ & 0.902$_{\pm 0.005}$ & 0.784$_{\pm 0.012}$ \\
    Tcn-ED & 0.853$_{\pm 0.015}$ & 0.848$_{\pm 0.050}$ & 0.843$_{\pm 0.011}$ & 0.707$_{\pm 0.082}$ & 0.850$_{\pm 0.021}$ & 0.758$_{\pm 0.011}$ & 0.862$_{\pm 0.019}$ & 0.881$_{\pm 0.036}$ & 0.846$_{\pm 0.003}$ & 0.658$_{\pm 0.090}$ & 0.868$_{\pm 0.023}$ & 0.763$_{\pm 0.008}$ \\
    DSVDD & 0.691$_{\pm 0.014}$ & 0.682$_{\pm 0.012}$ & 0.829$_{\pm 0.002}$ & 0.519$_{\pm 0.038}$ & 0.751$_{\pm 0.057}$ & 0.686$_{\pm 0.040}$ & 0.671$_{\pm 0.017}$ & 0.621$_{\pm 0.033}$ & 0.811$_{\pm 0.003}$ & 0.410$_{\pm 0.043}$ & 0.690$_{\pm 0.078}$ & 0.555$_{\pm 0.069}$ \\
    MSCRED & 0.766$_{\pm 0.036}$ & 0.628$_{\pm 0.031}$ & N/A   & 0.717$_{\pm 0.007}$ & 0.657$_{\pm 0.029}$ & 0.640$_{\pm 0.017}$ & 0.756$_{\pm 0.050}$ & 0.536$_{\pm 0.049}$ & N/A   & 0.644$_{\pm 0.016}$ & 0.659$_{\pm 0.062}$ & 0.604$_{\pm 0.023}$ \\
    Omni  & 0.810$_{\pm 0.044}$ & 0.954$_{\pm 0.006}$ & 0.845$_{\pm 0.012}$ & 0.738$_{\pm 0.018}$ & 0.867$_{\pm 0.018}$ & 0.811$_{\pm 0.027}$ & 0.789$_{\pm 0.063}$ & 0.928$_{\pm 0.011}$ & 0.841$_{\pm 0.008}$ & 0.714$_{\pm 0.023}$ & 0.914$_{\pm 0.015}$ & 0.780$_{\pm 0.054}$ \\
    USAD  & 0.595$_{\pm 0.033}$ & 0.744$_{\pm 0.006}$ & 0.835$_{\pm 0.000}$ & 0.666$_{\pm 0.049}$ & 0.733$_{\pm 0.041}$ & 0.663$_{\pm 0.009}$ & 0.510$_{\pm 0.035}$ & 0.658$_{\pm 0.006}$ & 0.808$_{\pm 0.000}$ & 0.611$_{\pm 0.044}$ & 0.713$_{\pm 0.065}$ & 0.541$_{\pm 0.038}$ \\
    GDN   & 0.801$_{\pm 0.034}$ & 0.939$_{\pm 0.015}$ & 0.846$_{\pm 0.025}$ & 0.640$_{\pm 0.052}$ & 0.795$_{\pm 0.022}$ & 0.783$_{\pm 0.010}$ & 0.779$_{\pm 0.042}$ & 0.959$_{\pm 0.016}$ & 0.885$_{\pm 0.036}$ & 0.659$_{\pm 0.043}$ & 0.728$_{\pm 0.024}$ & 0.789$_{\pm 0.010}$ \\
    NeuTraL & 0.627$_{\pm 0.047}$ & 0.770$_{\pm 0.050}$ & 0.862$_{\pm 0.023}$ & 0.681$_{\pm 0.085}$ & 0.534$_{\pm 0.035}$ & 0.739$_{\pm 0.075}$ & 0.592$_{\pm 0.045}$ & 0.746$_{\pm 0.067}$ & 0.850$_{\pm 0.026}$ & 0.616$_{\pm 0.111}$ & 0.490$_{\pm 0.045}$ & 0.729$_{\pm 0.111}$ \\
    TranAD & 0.899$_{\pm 0.020}$ & 0.761$_{\pm 0.001}$ & 0.831$_{\pm 0.009}$ & 0.755$_{\pm 0.010}$ & 0.730$_{\pm 0.114}$ & 0.776$_{\pm 0.008}$ & 0.915$_{\pm 0.018}$ & 0.662$_{\pm 0.003}$ & 0.810$_{\pm 0.007}$ &  
    \textbf{0.729$_{\pm 0.017}$} & 0.690$_{\pm 0.158}$ & 0.734$_{\pm 0.012}$ \\
    AnomTran & 0.679$_{\pm 0.134}$ & 0.657$_{\pm 0.097}$ & 0.845$_{\pm 0.006}$ & 0.704$_{\pm 0.025}$ & 0.836$_{\pm 0.067}$ & 0.817$_{\pm 0.012}$ & 0.627$_{\pm 0.166}$ & 0.608$_{\pm 0.139}$ & 0.822$_{\pm 0.010}$ & 0.639$_{\pm 0.019}$ & 0.868$_{\pm 0.066}$ & 0.814$_{\pm 0.029}$ \\
    \hline
    COUTA & \textbf{0.942$_{\pm 0.031}$} & \textbf{0.980$_{\pm 0.018}$} & \textbf{0.886$_{\pm 0.022}$} & \textbf{0.781$_{\pm 0.013}$} & \textbf{0.926$_{\pm 0.029}$} & \textbf{0.830$_{\pm 0.053}$} & \textbf{0.955$_{\pm 0.030}$} & \textbf{0.984$_{\pm 0.015}$} & \textbf{0.900$_{\pm 0.017}$} & 0.714$_{\pm 0.006}$ & \textbf{0.942$_{\pm 0.020}$} & \textbf{0.823$_{\pm 0.086}$} \\
		
			\hline
		\end{tabular}
	}
	\label{tab:effectiveness}%
\end{table*}%

\subsubsection{Evaluation Metrics and Computing Infrastructure} 


Many anomalies in time series data are consecutive, and they can be viewed as multiple anomaly segments. In many practical applications, an anomaly segment can be properly handled if detection models can trigger an alert at any timestamp within this segment. 
Therefore, the vast majority of previous studies \cite{su2019omni,li2021multivariate,tuli2022tranad,audibert2020usad,liu2022tkde,deng2021gdn,xu2022transformer} employ \textit{point-adjust} strategy in their experiment protocols. 
Specifically, the scores of observations in each anomaly segment are adjusted to the highest value within this segment, which simulates the above assumption (a single alert is sufficient to take action). Other points outside anomaly segments are treated as usual. 
For the sake of consistency with the current literature, we also employ this adjustment strategy before computing evaluation metrics.

Precision $P$ and recall $R$ of the anomaly class can directly indicate the costs and benefits of finding anomalies in real-world applications, which can intuitively reflect model performance. $F_1$ score is the harmonic mean of precision and recall, which takes both precision and recall into account. These detection models output continuous anomaly scores, but there is often no specific way to determine a decision threshold when calculating precision and recall. Therefore, following prior work in this research line \cite{su2019omni,li2021multivariate,tuli2022tranad,campos2021vldb,liu2022tkde,audibert2020usad}, we use the best $F_1$ score and the Area Under the Precision-Recall Curve (AUC-PR), considering simplicity and fairness. These two metrics can avoid possible bias brought by fixed thresholds or threshold calculation methods. The best $F_1$ score represents the optimal case, while AUC-PR is in an average case that is less optimal. Specifically, we enumerate all possible thresholds (i.e., scores of each timestamp) and compute corresponding precision and recall scores. The best $F_1$ can then be obtained, and AUC-PR is computed by employing the average precision score. 
In the following experiments, $F_1$, $P$, and $R$ denote the best $F_1$ and its corresponding precision and recall score. 
The above performance evaluation metrics range from 0 to 1, and a higher value indicates better performance. 


All the experiments are executed at a workstation with an Intel(R) Xeon(R) Silver 4210R CPU @ 2.4GHz, an NVIDIA TITAN RTX GPU, and 64 GB RAM.

\subsubsection{Parameter Settings and Implementations}
In COUTA, the temporal modeling network $\phi$ is with one hidden layer, and the kernel size uses 2. The projection head $\psi$ and the classification head $\psi_c$ are two-layer multi-layer perceptron networks with LeakyReLU activation. 
The hidden layer of $\phi$ and $\psi$ has 16 neural units by default, and the dimension of the feature space $\mathcal{F}$ is also 16. 
We respectively use 32 and 64 in complicated datasets \textit{ASD} and \textit{DSADS} to enhance the representation power of the neural network. Adam optimizer is employed, where the learning rate is set to $10^{-4}$.
The weight factor $\alpha$ of the supervised classification branch in the loss function uses 0.1 by default. 
The factor $\beta$ that controls the size of generated anomaly examples is set to 0.2. 
As for the competing methods, we use the default or recommended hyper-parameter settings in their respective original papers.

These anomaly detectors are implemented in Python. We use the implementations of OCSVM and ECOD from \texttt{pyod}, a python library of anomaly detection approaches. The source code of GOAD, DSVDD, USAD, GDN, NeuTraL, and TranAD are released by their original authors. 
In terms of LSTM-ED, Tcn-ED, MSCRED, and Omni, we use the implementations from an evaluation study \cite{garg2021evaluation}. 
The source code of COUTA is available at \url{https://github.com/xuhongzuo/couta}.


\begin{table}[t]
  \centering
  \caption{$F_1$ score and AUC-PR on four univariate datasets.}
  \scalebox{0.85}{
    \begin{tabular}
    {	p{0.2cm}	| 
       p{1.25cm}p{1.5cm}p{1.5cm}
       p{1.6cm}p{1.6cm} 
    }
    \hline
    & \textbf{Methods} & \textbf{UCR-Fault} & \textbf{UCR-Gait} & \textbf{UCR-ECG-s} & \textbf{UCR-ECG-v} \\
    \hline
    
    
    \multirow{11}[0]{*}
    {\begin{sideways}\textbf{\boldmath$F_1$\unboldmath}\end{sideways}} & ARMA & \textbf{1.000$_{\pm 0.000}$} & \textbf{0.992$_{\pm 0.000}$} & 0.906$_{\pm 0.000}$ & 0.824$_{\pm 0.000}$ \\
          
    & LSTM-Pr & \textbf{1.000$_{\pm 0.000}$} & 0.865$_{\pm 0.052}$ & 0.917$_{\pm 0.014}$ & 0.919$_{\pm 0.099}$ \\
    & LSTM-ED & 0.685$_{\pm 0.103}$ & 0.815$_{\pm 0.036}$ & 0.760$_{\pm 0.024}$ & 0.723$_{\pm 0.062}$ \\
    & Tcn-ED & 0.994$_{\pm 0.010}$ & 0.912$_{\pm 0.161}$ & 0.867$_{\pm 0.059}$ & 0.679$_{\pm 0.228}$ \\
    & DSVDD & 0.846$_{\pm 0.028}$ & 0.869$_{\pm 0.011}$ & 0.850$_{\pm 0.057}$ & 0.783$_{\pm 0.074}$ \\
    & Omni  & 0.316$_{\pm 0.094}$ & 0.426$_{\pm 0.138}$ & 0.759$_{\pm 0.013}$ & 0.828$_{\pm 0.068}$ \\
    & USAD  & 0.519$_{\pm 0.072}$ & 0.567$_{\pm 0.103}$ & 0.359$_{\pm 0.014}$ & 0.510$_{\pm 0.092}$ \\
    & NeuTraL & 1.000$_{\pm 0.000}$ & 0.804$_{\pm 0.032}$ & 0.711$_{\pm 0.028}$ & 0.790$_{\pm 0.071}$ \\
    & TranAD & 0.877$_{\pm 0.117}$ & 0.916$_{\pm 0.040}$ & 0.779$_{\pm 0.029}$ & 0.814$_{\pm 0.001}$ \\
    & AnomTran & 0.786$_{\pm 0.212}$ & 0.781$_{\pm 0.181}$ & 0.863$_{\pm 0.094}$ & 0.827$_{\pm 0.187}$ \\
    \hline
    & COUTA & \textbf{1.000$_{\pm 0.000}$} & 0.976$_{\pm 0.034}$ & \textbf{0.924$_{\pm 0.038}$} & \textbf{0.968$_{\pm 0.014}$} \\
    \hline
    
    \multirow{11}[0]{*}{\begin{sideways}\textbf{AUC-PR}\end{sideways}} & ARMA & \textbf{1.000$_{\pm 0.000}$} & \textbf{0.984$_{\pm 0.000}$} & 0.835$_{\pm 0.000}$ & 0.752$_{\pm 0.000}$ \\
          
    & LSTM-Pr & \textbf{1.000$_{\pm 0.000}$} & 0.777$_{\pm 0.072}$ & 0.855$_{\pm 0.021}$ & 0.899$_{\pm 0.123}$ \\
    & LSTM-ED & 0.559$_{\pm 0.114}$ & 0.698$_{\pm 0.046}$ & 0.646$_{\pm 0.035}$ & 0.627$_{\pm 0.077}$ \\
    & Tcn-ED & 0.989$_{\pm 0.018}$ & 0.886$_{\pm 0.197}$ & 0.780$_{\pm 0.084}$ & 0.566$_{\pm 0.289}$ \\
    & DSVDD & 0.791$_{\pm 0.028}$ & 0.784$_{\pm 0.015}$ & 0.763$_{\pm 0.074}$ & 0.702$_{\pm 0.081}$ \\
    & Omni  & 0.216$_{\pm 0.078}$ & 0.313$_{\pm 0.127}$ & 0.636$_{\pm 0.019}$ & 0.769$_{\pm 0.081}$ \\
    & USAD  & 0.460$_{\pm 0.068}$ & 0.463$_{\pm 0.094}$ & 0.266$_{\pm 0.015}$ & 0.389$_{\pm 0.077}$ \\
    & NeuTraL & 1.000$_{\pm 0.000}$ & 0.717$_{\pm 0.037}$ & 0.622$_{\pm 0.030}$ & 0.701$_{\pm 0.099}$ \\
    & TranAD & 0.801$_{\pm 0.172}$ & 0.849$_{\pm 0.068}$ & 0.689$_{\pm 0.030}$ & 0.758$_{\pm 0.001}$ \\
    & AnomTran & 0.757$_{\pm 0.249}$ & 0.703$_{\pm 0.211}$ & 0.779$_{\pm 0.141}$ & 0.758$_{\pm 0.224}$ \\
    \hline
    & COUTA & \textbf{1.000$_{\pm 0.000}$} & 0.956$_{\pm 0.060}$ & \textbf{0.871$_{\pm 0.060}$} & \textbf{0.941$_{\pm 0.025}$} \\
    \hline
    \end{tabular}%
    }
  \label{tab:effectiveness2}%
\end{table}%

\subsection{Effectiveness in Real-World Datasets}\label{sec:effectiveness}
This experiment evaluates the effectiveness of COUTA. We perform COUTA and its competing methods on ten real-world time series datasets. These models are trained on training sets, and we report their detection performance on testing sets. Ground-truth labels in testing sets are strictly unknown in the training stage. 

Table \ref{tab:effectiveness} and Table \ref{tab:effectiveness2} respectively illustrate the $F_1$ score and AUC-PR of anomaly detectors on six multivariate datasets and four univariate datasets. 
ARMA cannot obtain results in three days on SMD, SWaT, and DSADS. 
MSCRED runs out of memory on the large-scale dataset \textit{SWaT} due to the high computational cost of the deep convolutional network structure used in MSCRED. It is noteworthy that OCSVM, GOAD, ECOD, and DAGMM are not originally designed for time series data. MSCRED and GDN learn interactions between different varieties, i.e., they cannot perform single-dimensional data. Thus, we report COUTA and the remaining nine competitors in Table \ref{tab:effectiveness2}. 
COUTA obtains the best $F_1$ performance on all datasets. In terms of AUC-PR, COUTA outperforms all of its state-of-the-art competing methods on nine datasets with a 0.015 disparity to the best performer on the remaining dataset. 
Averagely, COUTA achieves 11\% - 49\% $F_1$ improvement and 11\% - 68 \% AUC-PR enhancement over various competitors.

According to the comparison results, COUTA successfully achieves state-of-the-art performance by addressing two key limitations in the current one-class learning pipeline. The superiority of COUTA can be attributed to the synergy of our two novel one-class calibration components, which achieves contamination-tolerant, anomaly-informed learning of data normality. 
The canonical one-class learning process used in these competing methods suffers from the anomaly contamination problem, and they may also learn an inaccurate range of normal behaviors due to the lack of guidance about the anomaly class. 
It is noteworthy that, on dataset \textit{SWaT}, a pure training set with only normal observations is ensured (water treatment attack is only launched in the testing set), and thus competing methods can obtain relatively good performance. 
AnomTran, TranAD, and GDN achieve better performance compared to other competitors. AnomTran and TranAD employ the advanced Transformer structure and several training tricks, and GDN models inter-variate correlations with the help of the powerful capability of graph neural networks in exploring high-order interactions among graph nodes. However, TranAD may suffer from the overfitting problem because its learning process contains several complicated components, and thus it fails to produce effective detection results on simple datasets like \textit{SMD} and \textit{SWaT}. In contrast, detectors with plain encoder-decoder structures (e.g., LSTM-ED) obtain better performance.

\subsection{Generalization Ability to Different Types of Time Series Anomalies}\label{sec:generalization}

\begin{figure}[t]
	\centering
	\includegraphics[width=0.49\textwidth]{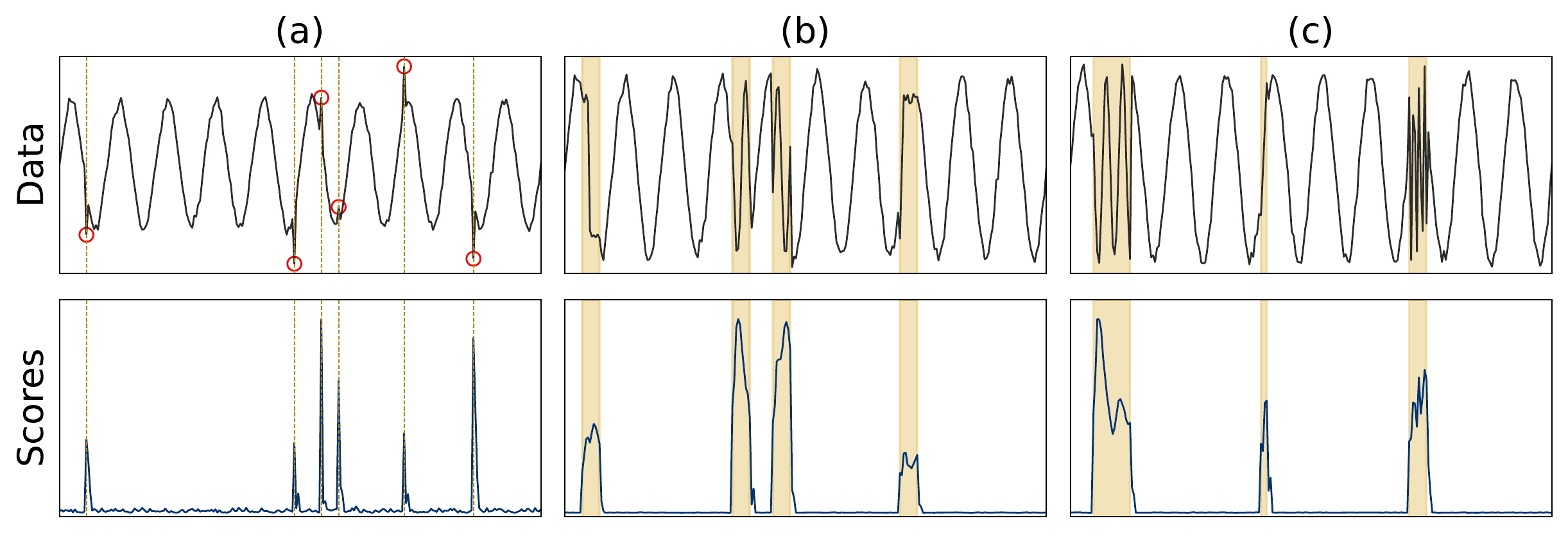}
	\caption{Performance of COUTA in handling (a) point-wise (global and contextual) anomalies, (b) pattern-wise (seasonal and shapelet) anomalies, and (c) anomalies with different lengths. 
	}
	\label{fig:case}
\end{figure}

We investigate whether COUTA can handle different anomaly types in time series data. Following a fine-grained taxonomy of time series anomalies in \cite{lai2021revisiting}, three synthetic datasets are created. 
These datasets contain 1000 observations, which are described in two dimensions. The first 400 points are used for training, and the remaining data are treated as testing sets. 
We demonstrate the testing data of three cases in Fig. \ref{fig:case}, where (a), (b), and (c) contain point-wise anomalies, pattern-wise anomalies, and pattern-wise anomalies with different lengths, respectively. Point-wise anomalies include both global and contextual forms, and pattern-wise anomalies are caused by basic shapelet and seasonality changes. 

The bottom panel of Fig. \ref{fig:case} shows the anomaly scores produced by our method COUTA. 
COUTA successfully identifies all of these anomaly cases with distinguishably higher scores on true anomalies and consistently lower scores on normal moments. We use three pre-defined fixed operations in creating native anomaly examples. These native anomaly examples are used to calibrate the normality boundary. COUTA can still generalize to different anomaly types even if some anomaly types may not strictly correspond to the generated native anomalies.
It is mainly because COUTA is essentially a one-class classification model, and by definition, it can alarm all the observations that deviate from the learned normality according to one-class distances. The data normality is modeled based on the majority of the training data instead of these created native anomalies. Therefore, the generalization ability of our method COUTA is not limited by the pre-defined operations in NAC, i.e., COUTA can also identify novel anomalies that are not covered in the NAC process.


\begin{figure}[t]
	\centering
	\includegraphics[width=0.48\textwidth]{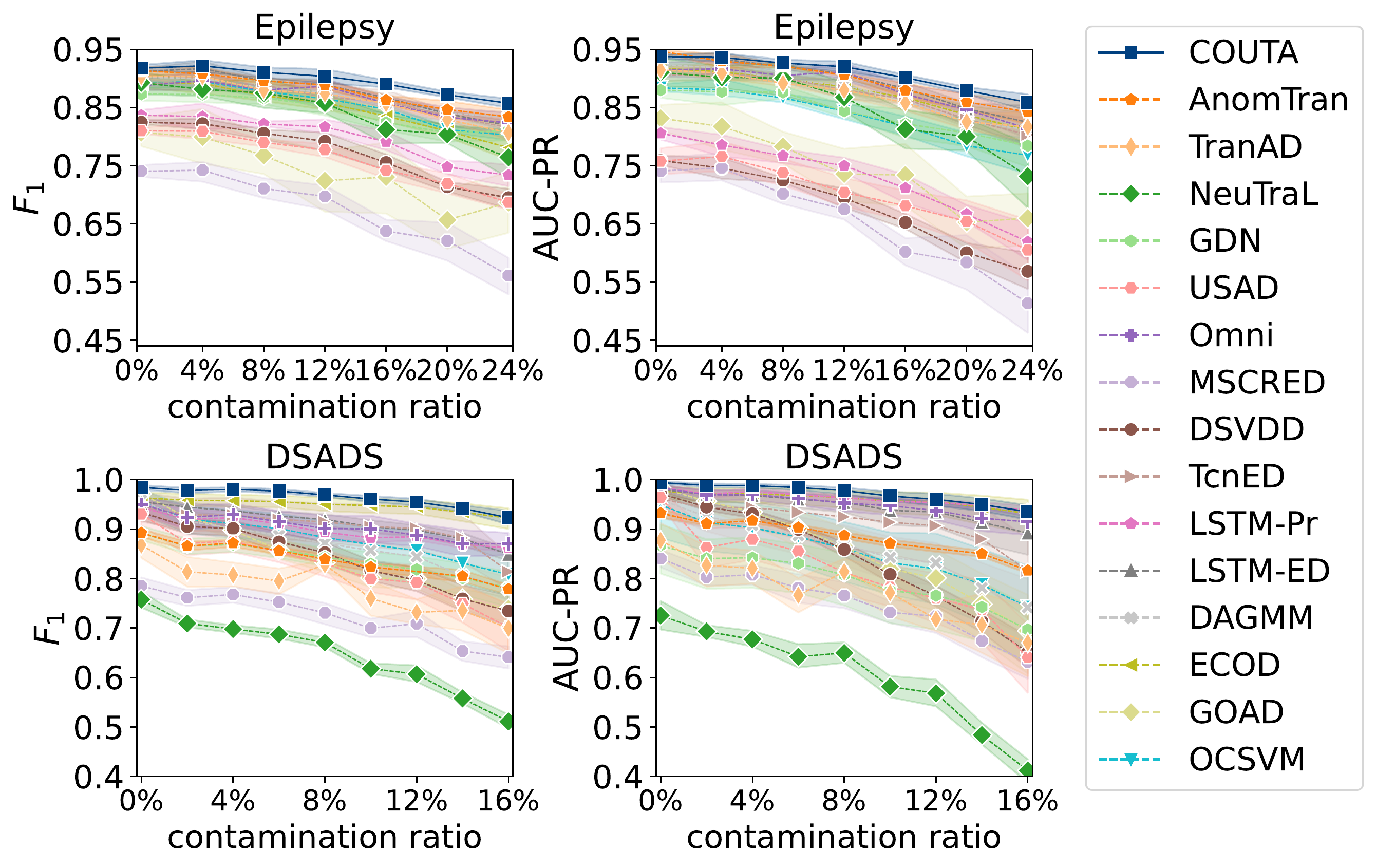}
	\caption{Robustness w.r.t. anomaly contamination. }
	\label{fig:robustness}
\end{figure}


\subsection{Robustness w.r.t. Anomaly Contamination}


This experiment is to quantitatively measure the interference of anomaly contamination to time series anomaly detectors, that is, we test the robustness of each anomaly detector w.r.t. different anomaly contamination ratios in the training set. Due to the continuity of time series data, we cannot directly remove or inject anomalies like other robustness testing experiments in prior work on tabular data \cite{xu2023rosas,xu2023deep}. It is hard to precisely adjust ratios of abnormal timestamps in the training set. Therefore, we employ \textit{Epilepsy} and \textit{DSADS} in this experiment because these datasets are collections of divided small sequences with sequence-level labels. We treat small sequences as data objects and generate a group of datasets with different anomaly contamination levels of the training set. We first randomly remove anomaly sequences in the training set to meet the requirements of specific contamination ratios. The removed anomalies are then added to the testing set.
The original sequence-level anomaly contamination rate in \textit{Epilepsy} and \textit{DSADS} are 24\% and 16\%, respectively. A wide range of contamination levels is used for each dataset by starting from a pure training set and taking 4\% as the increasing step. 

Fig. \ref{fig:robustness} presents the $F_1$ score and the AUC-PR performance of COUTA and its competing methods on datasets with different anomaly contamination ratios. 
The performance of all anomaly detectors downgrades with the increase of anomaly contamination. COUTA shows better robustness compared to its contenders, especially on datasets with a large contamination rate. It owes to the novel one-class classification loss function, which successfully masks these noisy data via uncertainty modeling-based adaptive penalty. 
By contrast, these competing methods simply view these hidden anomalies as normal data, and the learned normality model may mistakenly overfit these noises. This experiment validates the contribution of the proposed uncertainty modeling-based calibration method. This experiment also shows superior applicability of COUTA in some real-world applications that are with complicated noisy circumstances.

\subsection{Scalability Test}
This experiment investigates the scalability of COUTA compared to its competing methods. Time efficiency w.r.t. both time series length $T$ and dimensionality $D$ are recorded. As for the scalability test w.r.t. dimensionality, a group of seven time-series datasets with a fixed length (i.e., 2,000) and various dimensions (i.e., from 8 to 512 with 2 as the magnification factor) is generated. We synthesize another group of nine datasets containing different lengths (i.e., range from 1,000 to 256,000) and a fixed dimension (i.e., 8) for the scale-up test w.r.t. length. As these deep anomaly detectors are deployed with different numbers of training epochs, we report the execution time of one training epoch by taking 128 as a unified size of training mini-batches. 

\begin{figure}[t]
	\centering
	\includegraphics[width=0.49\textwidth]{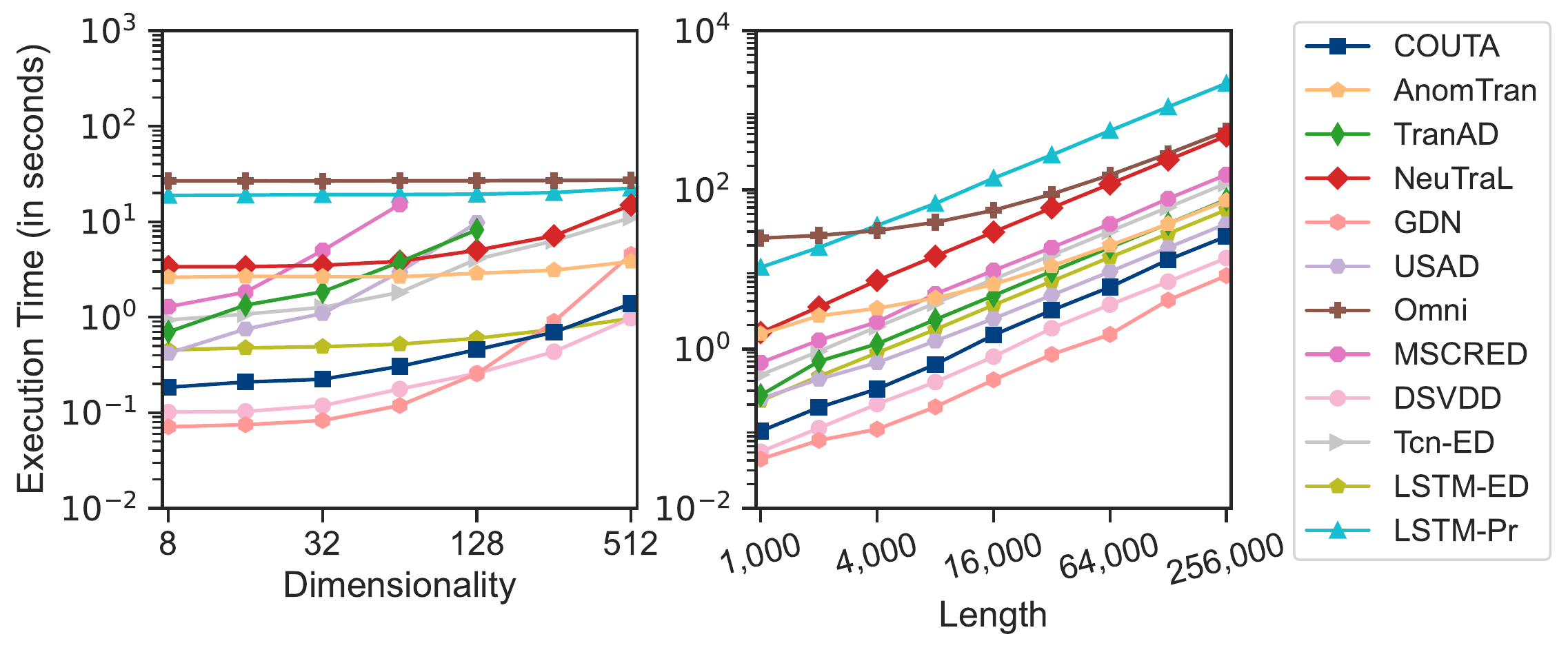}
	\caption{Scalability test results w.r.t. dimensionality and length of time series. TranAD, USAD, and MSCRED run out of memory when handling time series data with high dimensionality. }
	\label{fig:scalability}
\end{figure}

Fig. \ref{fig:scalability} presents the execution time of COUTA and its ten competing state-of-the-art methods on time series datasets with various sizes. Note that this experiment excludes three general anomaly detectors (i.e., OCSVM, ECOD, and GOAD) that are not originally designed for time series data. COUTA has good scalability compared to most of these existing methods, which shows its potential capability of being applied in practical scenarios where time series data are with large volumes and high dimensions.


\subsection{Sensitivity Test}

This experiment shows the impact of different parameter settings on detection performance. First, we only use a single kind of perturbation ($\delta_{\text{I}}$, $\delta_{\text{II}}$, or $\delta_{\text{III}}$) to solely generate point, contextual, or collective anomalies in NAC. Besides, we investigate several key hyper-parameters of COUTA including $\alpha$, $\beta$, $H$, and $l$. $\alpha$ is the weight of loss $\mathcal{L}_{\text{NAC}}$ in Equation (\ref{eqn:loss0}), $\beta$ is the ratio that controls the size of generated anomaly examples, $H$ is the dimensionality of the feature space $\mathcal{F}$, and $l$ is the sliding window length. Each parameter is sampled from a wide range.

\begin{figure}[t]
	\centering
	\includegraphics[width=0.48\textwidth]{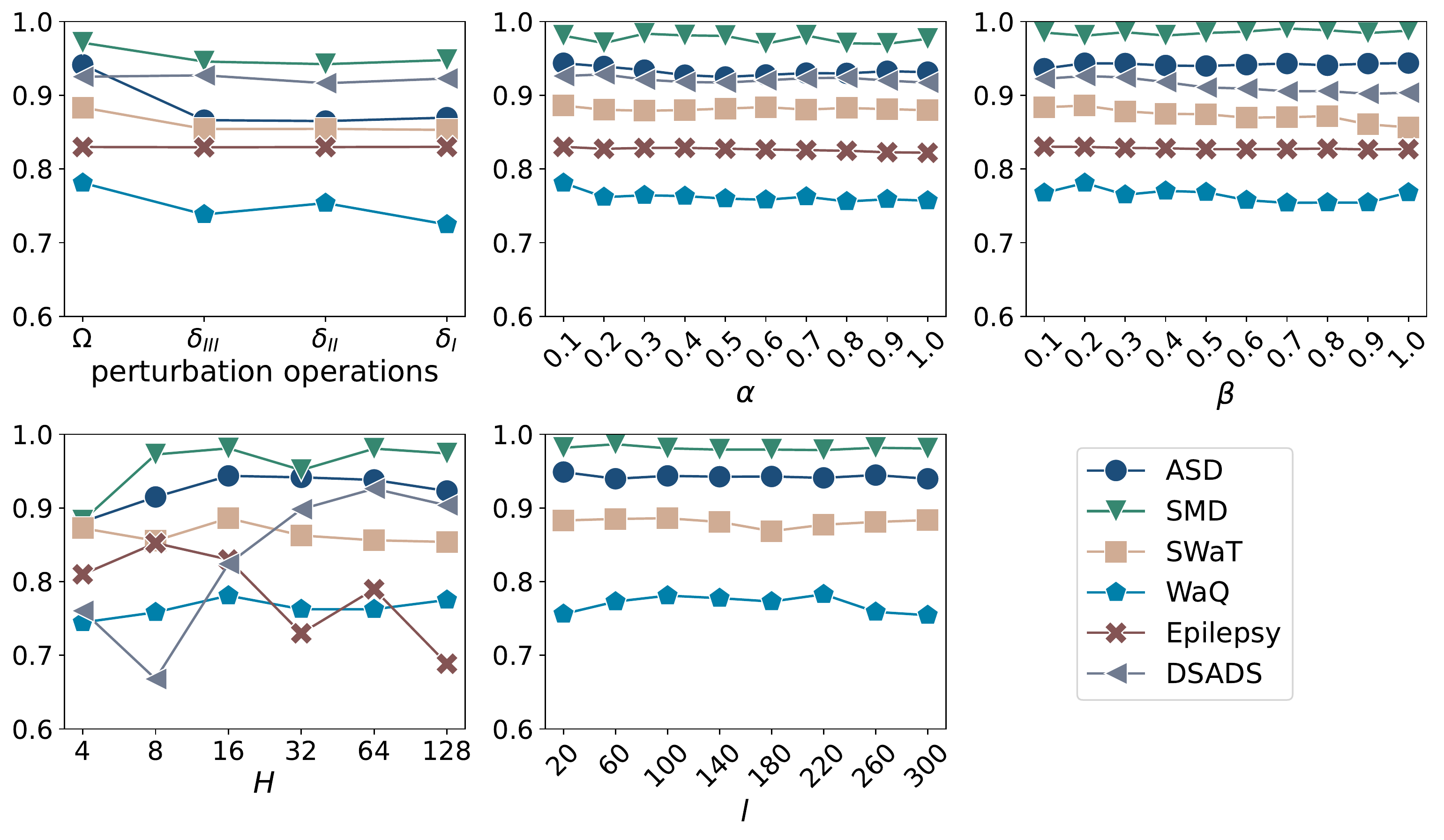}
	\caption{Sensitivity test results ($F_1$ performance w.r.t. different parameter settings). \textit{Epilepsy} and \textit{DSADS} have default lengths of sub-sequences, and thus the detection performance w.r.t. $l$ on these two datasets is omitted. }
	\label{fig:para}
\end{figure}

Fig. \ref{fig:para} shows the $F_1$ performance of COUTA by taking different parameter settings, and AUC-PR performance also shows the same trend, which is omitted here.  
COUTA shows better performance when a full perturbation operation pool $\Omega$ is employed in NAC (especially on the \textit{ASD} dataset). A single type of generated anomaly example may provide fragmentary knowledge about the anomaly class, thus leading to less optimal performance. 
In terms of $\alpha$, $\beta$, and $l$, these parameters do not largely influence the performance, and COUTA performs stably with different hyper-parameters. Some parameter selection methods can be employed. 
There might be some unreliable generated anomaly examples that fall into the normal distribution, and thus we use $\alpha\!=\!0.1$, $\beta\!=\!0.2$ by default. In terms of $l$, we employ a frequently-used sliding window length (i.e., $l=100$).
COUTA shows fluctuation trends w.r.t. the dimensionality of the feature space $H$. It might be because these time series datasets are with various numbers of dimensions. 
Generally, a larger $H$ might be preferable when handling high-dimensional time series datasets (e.g., \textit{DSADS} with 47 dimensions) because the representation capability can be ensured, while low-dimensional datasets can be well processed by using smaller $H$ (e.g., \textit{Epilepsy} with 5 dimensions).

\subsection{Ablation Study}
\begin{figure}[t]
	\centering
	\includegraphics[width=0.5\textwidth]{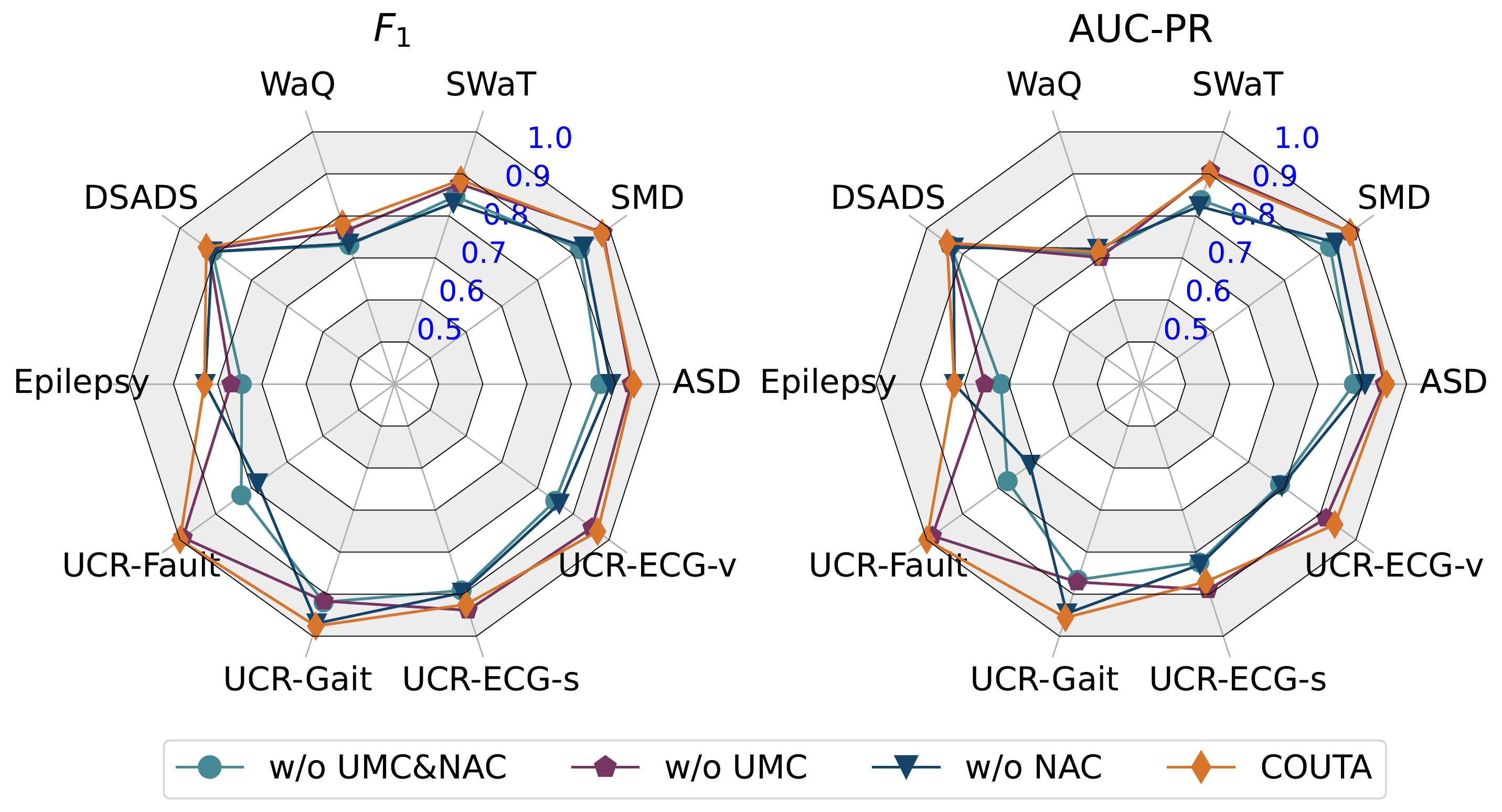}
	\caption{Ablation study results.}
	\label{fig:ablation}
\end{figure}

This experiment further validates the contribution of two key components in COUTA.
Three ablated variants are used. 
Two calibration methods, i.e., uncertainty modeling-based calibration and native anomaly-based calibration, are respectively removed from COUTA in two ablated variants, \textbf{w/o} UMC and \textbf{w/o} NAC. These two components are simultaneously excluded in \textbf{w/o} UMC\&NAC. The remaining parts of these ablated versions remain the same as COUTA. 
We report the $F_1$ and AUC-PR performance of standard COUTA and its three ablated versions in Fig. \ref{fig:ablation}. 
Based on the comparison of COUTA and its variants, the superiority of COUTA can verify the significant contribution of two calibration methods on one-class classification. 
COUTA outperforms \textbf{w/o} UMC, \textbf{w/o}, and \textbf{w/o} UMC\&NAC by 8\%, 2\%, and 7\%, respectively. 
Particularly, UMC conduces to 11\% improvement in $F_1$ score and 14\% gain in AUC-PR on the \textit{Epilepsy} dataset, where the training set is severely contaminated by unknown anomalies, and NAC can bring approximate 8 \% enhancement in both $F_1$ and AUC-PR on the \textit{ASD} dataset.


\section{Conclusions}
This paper introduces COUTA, an unsupervised time series anomaly detection method based on calibrated one-class classification. 
We address two key challenges in the current one-class learning pipeline, i.e., the presence of anomaly contamination and the absence of knowledge about anomalies. 
COUTA achieves this goal through two novel calibration methods -- uncertainty modeling-based calibration (UMC) and native anomaly-based calibration (NAC). In UMC, we obtain model uncertainty by imposing a prior distribution to the one-class distance value, and a theoretically motivated novel learning objective is devised to restrain noisy data that are with high uncertainty, while simultaneously encouraging confident predictions to ensure effective learning of hard normal samples. 
In NAC, we design tailored data perturbation operations to produce native anomaly examples based on original time series data, which provides one-class classification with valuable knowledge about primeval anomalous behaviors.
These calibration methods enable COUTA to learn data normality in a noise-tolerant, anomaly-informed manner.
Extensive experiments show that COUTA achieves state-of-the-art performance in time series anomaly detection by substantially outperforming sixteen competitors. We also validate several desired properties of COUTA, including outstanding generalization ability to different anomaly types, superior robustness to anomaly contamination, and good scalability. 

Similar to many one-class classification methods, our approach also assumes that normal data share similarities and belong to one prototype. Our approach is adept at handling time series data that are seasonal and have many repetitive patterns, whereas our approach may fail when there are concept drifts happening in time series.  

In the future, we plan to investigate new strategies to estimate the mean and variance of the prior one-class distance distribution. Also, in producing native anomalies, tailored perturbation methods can be extended to a heuristic generation process.



\ifCLASSOPTIONcompsoc
\section*{Acknowledgments}
\else
\section*{Acknowledgment}
\fi

The work of Hongzuo Xu, Yijie Wang, Songlei Jian, Qing Liao, and Yongjun Wang were supported 
in part by the National Key R\&D Program of China under
Grant 2022ZD0115302, 
in part by the National Natural Science Foundation of China under Grant 62002371, 62076079, U19A2067, 61379052, and 61472439, 
in part by the Science Foundation of Ministry of Education of China under Grant 2018A02002, 
in part by the Natural Science Foundation for Distinguished Young Scholars of Hunan Province under Grant 14JJ1026, 
in part by the Foundation of National University of Defense Technology under Grant ZK21-17. 
The work of Guansong Pang was supported in part by the Singapore Ministry of Education (MOE) Academic Research Fund (AcRF) Tier 1 under Grant 21SISSMU031.

We thank Prof. Eamonn Keogh and his students for creating the UCR time series anomaly archive that has been used in our experiment. We also thank the referees for their comments, which helped improve this paper considerably.

\ifCLASSOPTIONcaptionsoff
\newpage
\fi


\bibliographystyle{IEEEtran}
\bibliography{ref}

%

\begin{IEEEbiography}[{\includegraphics[width=1in,height=1.25in,clip,keepaspectratio]{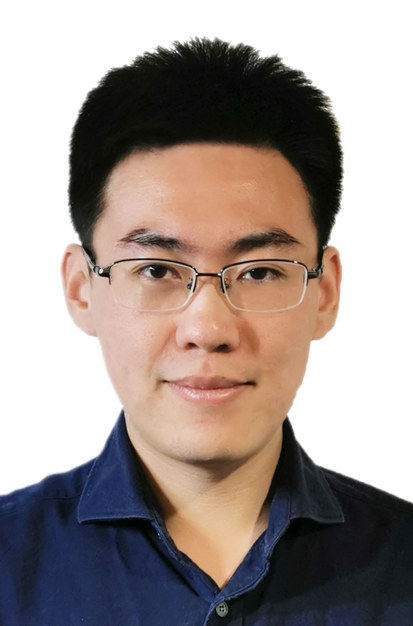}}]{Hongzuo Xu}
received the Ph.D degree in Computer Science and Technology from the National University of Defense Technology in 2023. 
His research interests include anomaly detection and its applications, with first-authored publications in ICML, WWW, AAAI, ICDM, CIKM, and \textsc{IEEE Transactions on Knowledge and Data Engineering}.  
\end{IEEEbiography}

\begin{IEEEbiography}[{\includegraphics[width=1in,height=1.25in,clip,keepaspectratio]{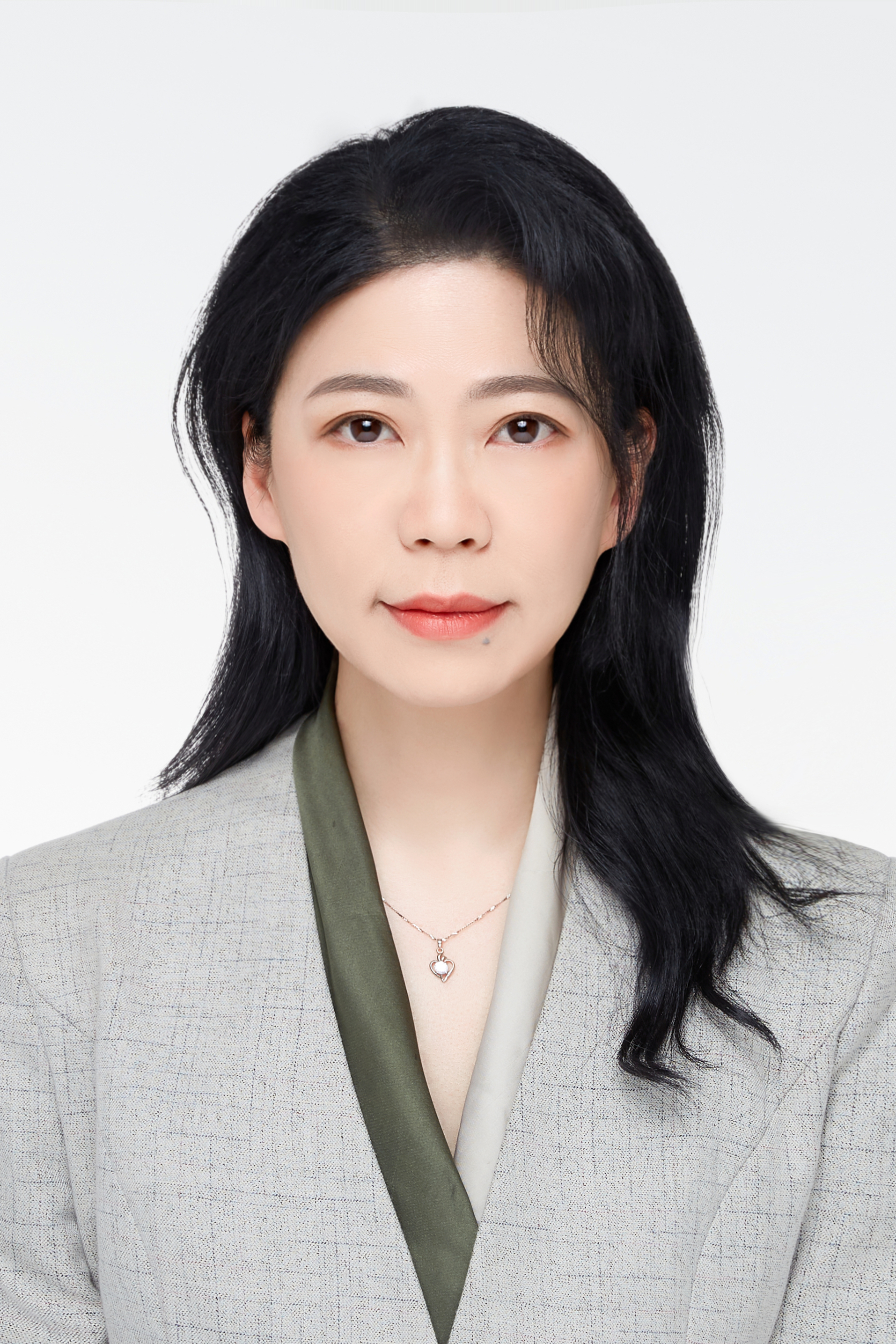}}]{Yijie Wang}
	received the PhD degree in computer science and technology from the National University of Defense Technology in 1998. 
	She was awarded the prize of National Excellent Doctoral Dissertation by Ministry of Education of PR China.
	She is currently a Full Professor with the National Key Laboratory of Parallel and Distributed Computing, National University of Defense Technology. Her research interests include	big data analysis, artificial intelligence and parallel and distributed processing.
\end{IEEEbiography}

\begin{IEEEbiography}[{\includegraphics[width=1in,height=1.25in,clip,keepaspectratio]{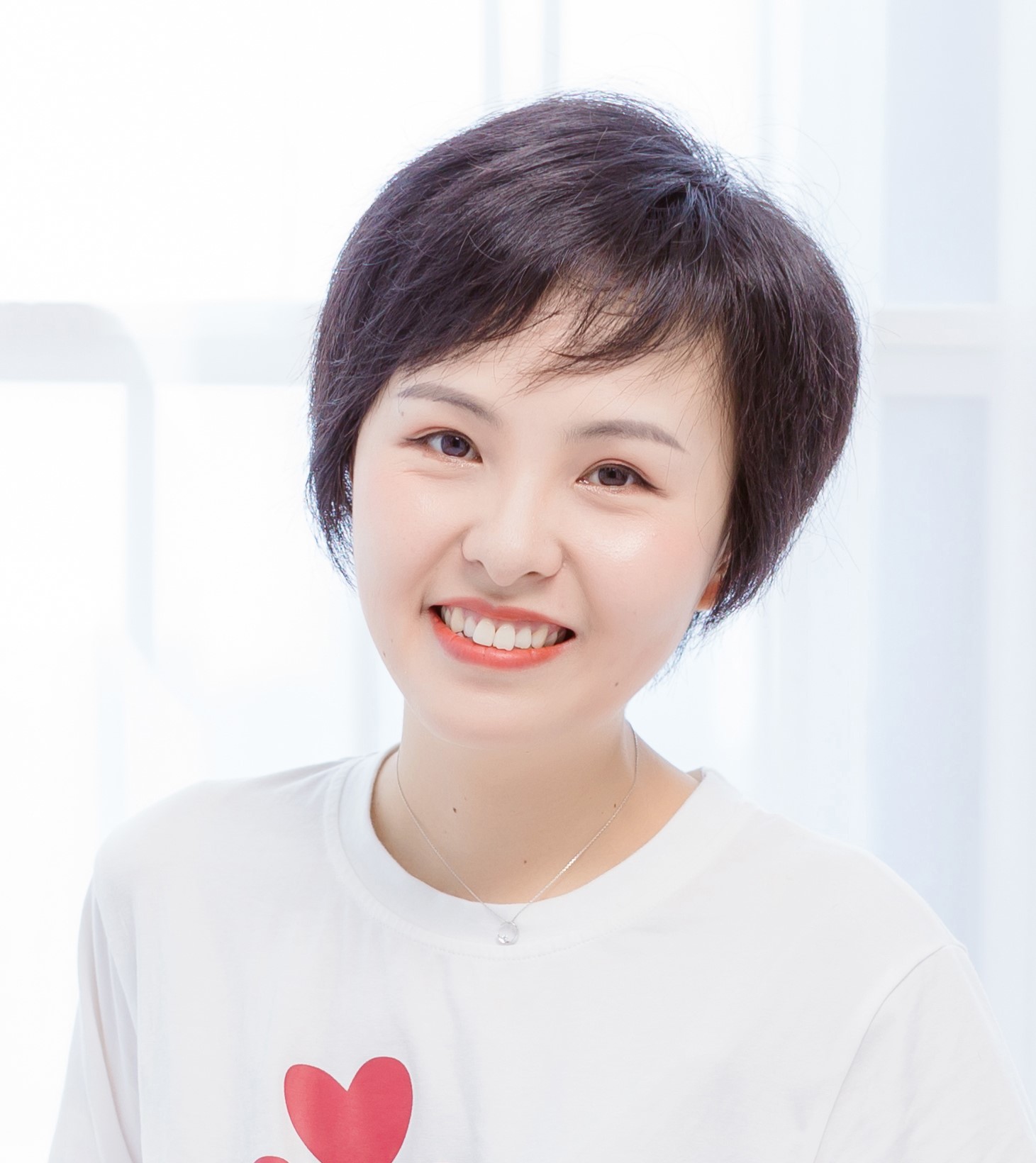}}]{Songlei Jian}
	received the B.Sc. degree and Ph.D. degree in computer science from the College of Computer, National University of Defense Technology, Changsha, China, in 2013 and 2019, respectively. She is currently an Associate Research Fellow with the College of Computer, National University of Defense Technology. Her research interests include representation learning, graph learning, multimodal learning and anomaly detection.
\end{IEEEbiography}

\begin{IEEEbiography}[{\includegraphics[width=1in,height=1.25in,clip,keepaspectratio]{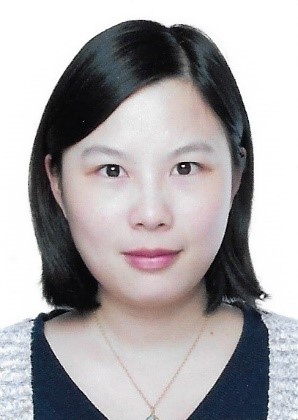}}]{Qing Liao}
received the Ph.D degree in computer science and engineering from the Department of Computer Science and Engineering of the Hong Kong University of Science and Technology, in 2016 supervised by professor Qian Zhang. She is currently a professor with the School of Computer Science and Technology, Harbin Institute of Technology (Shenzhen), China. Her research interests include artificial intelligence and data mining.
\end{IEEEbiography}

\begin{IEEEbiography}[{\includegraphics[width=1in,height=1.25in,clip,keepaspectratio]{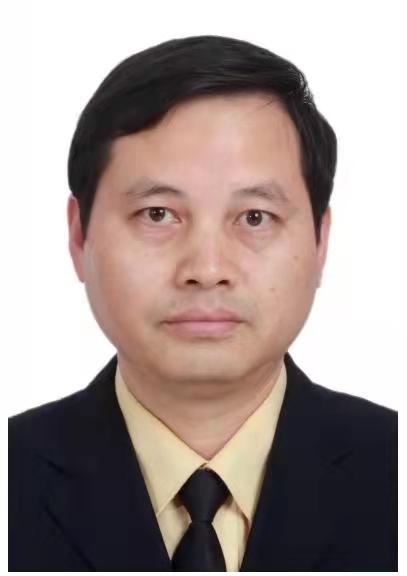}}]{Yongjun Wang}
received the Ph.D. degree in computer architecture from the National University of Defense Technology, China, in 1998. He is currently a Full Professor with the College of Computer, National University of Defense Technology, Changsha, China. His research interests include network security and system security.
\end{IEEEbiography}

\begin{IEEEbiography}[{\includegraphics[width=1in,height=1.25in,clip,keepaspectratio]{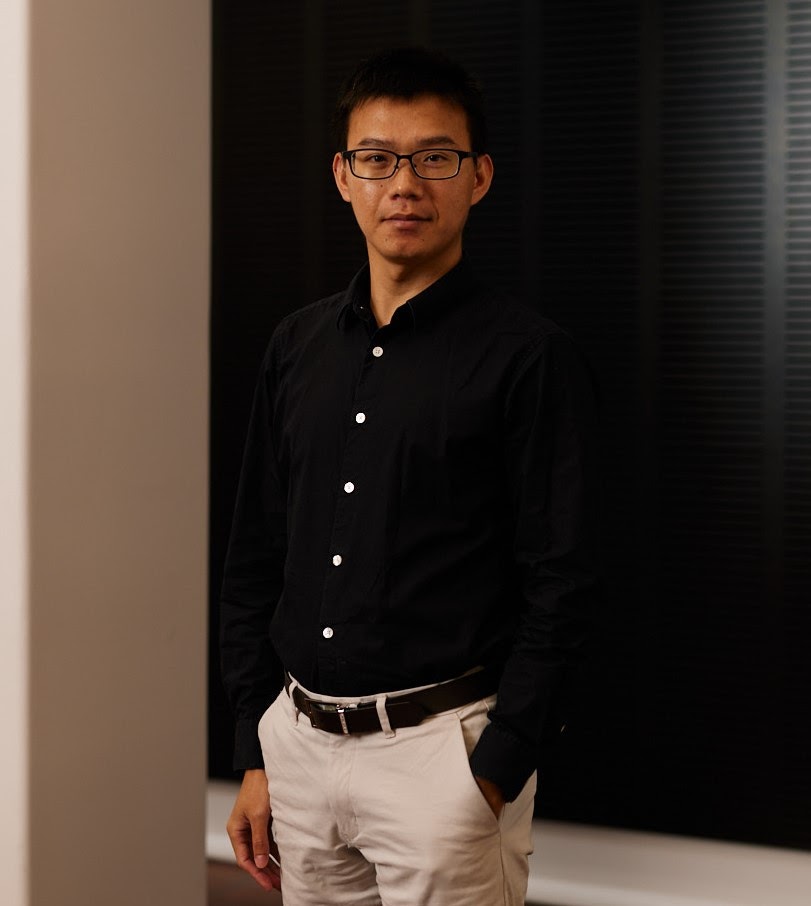}}]{Guansong Pang} is a tenure-track Assistant Professor of Computer Science in the School of Computing and Information Systems at Singapore Management University (SMU), Singapore. Before joining SMU, he was a Research Fellow with the Australian Institute for Machine Learning (AIML) at The University of Adelaide, Australia. He received a PhD degree from University of Technology Sydney in 2019. His research investigates novel data mining and machine learning techniques and their applications.
\end{IEEEbiography}







\end{document}